%% file: main.tex
\documentclass[10pt,twocolumn,letterpaper]{article}

\usepackage{cvpr}              % To produce the CAMERA-READY version
\input{preamble}

\definecolor{cvprblue}{rgb}{0.21,0.49,0.74}
\usepackage[pagebackref,breaklinks,colorlinks,allcolors=cvprblue]{hyperref}

\def\paperID{6129} % *** Enter the Paper ID here
\def\confName{CVPR}
\def\confYear{2026}

\title{\method: Tool-Augmented Multi Agents for Iterative 3D Object Arrangement}

\author{
Zhengfei Kuang\textsuperscript{1}\thanks{This work was done when Zhengfei Kuang was an intern at Google.}, Rui Lin \textsuperscript{2}, Long Zhao\textsuperscript{2}, Gordon Wetzstein\textsuperscript{1}, Saining Xie\textsuperscript{2,3}, Sanghyun Woo\textsuperscript{2}\\
\textsuperscript{1}Stanford University\qquad 
\textsuperscript{2}Google\qquad 
\textsuperscript{3}New York University\\
{\tt\small \{zhengfei,gordonwz\}@stanford.edu \{linrui,shwoo\}@google.com}\\
{\tt\small gary.zhao9012@gmail.com \qquad   saining.xie@nyu.edu}\\
{\tt\small \url{vulcan-3d.github.io}}
}

\begin{document}
\maketitle

\input{sec/teaser}
\input{sec/0_abstract}
\input{sec/1_intro}

\input{sec/2_related_work}

\input{sec/3_methods}

\input{sec/4_experiments}
\input{sec/5_discussion}

{
    \small
    \bibliographystyle{ieeenat_fullname}
    \bibliography{main}
}

\clearpage
\input{sec/X_suppl}

\end{document}

%% file: preamble.tex
\usepackage[dvipsnames]{xcolor}
\usepackage{overpic}
\usepackage{enumitem} %< control spacing in itemize/enumerate/...
\usepackage{overpic} %< add raw math symbols to figures
\usepackage{color}
\usepackage{array}
\usepackage{bm}
\usepackage{multirow}
\usepackage{multicol}
\usepackage{graphicx}
\usepackage{amsmath}
\usepackage{amssymb}
\usepackage{booktabs}
\usepackage{cuted}
\usepackage{capt-of}
\usepackage{algorithm2e}

\newcommand{\ignorethis } [1] {}
\definecolor{turquoise}{cmyk}{0.65,0,0.1,0.3}
\definecolor{purple}{rgb}{0.65,0,0.65}
\definecolor{dark_green}{rgb}{0, 0.5, 0}
\definecolor{orange}{rgb}{0.8, 0.6, 0.2}
\definecolor{red}{rgb}{0.8, 0.2, 0.2}
\definecolor{darkred}{rgb}{0.6, 0.1, 0.05}
\definecolor{blueish}{rgb}{0.0, 0.3, .6}
\definecolor{light_gray}{rgb}{0.7, 0.7, .7}
\definecolor{pink}{rgb}{1, 0, 1}
\definecolor{greyblue}{rgb}{0.25, 0.25, 1}

\definecolor{rone}{HTML}{66c5cc}
\definecolor{rtwo}{HTML}{f6cf71}
\definecolor{rthree}{HTML}{f89c74}

\definecolor{color1}{rgb}{0.36470588235, 0.7294117647, 0.43921568627}
\definecolor{color2}{rgb}{0.88235294117, 0.27450980392, 0.2431372549}
\definecolor{color3}{rgb}{0.58823529411, 0.46666666666, 0.4}
\definecolor{color4}{rgb}{0.30196078431, 0.62352941176, 0.88235294117}
\definecolor{color5}{rgb}{0.8862745098, 0.66274509803, 0.22745098039}

\newcommand{\method}{\textbf{VULCAN}\xspace}
\newcommand{\comment}[1]{}

%% file: sec/teaser.tex
\begin{strip}\centering
\includegraphics[width=\textwidth]{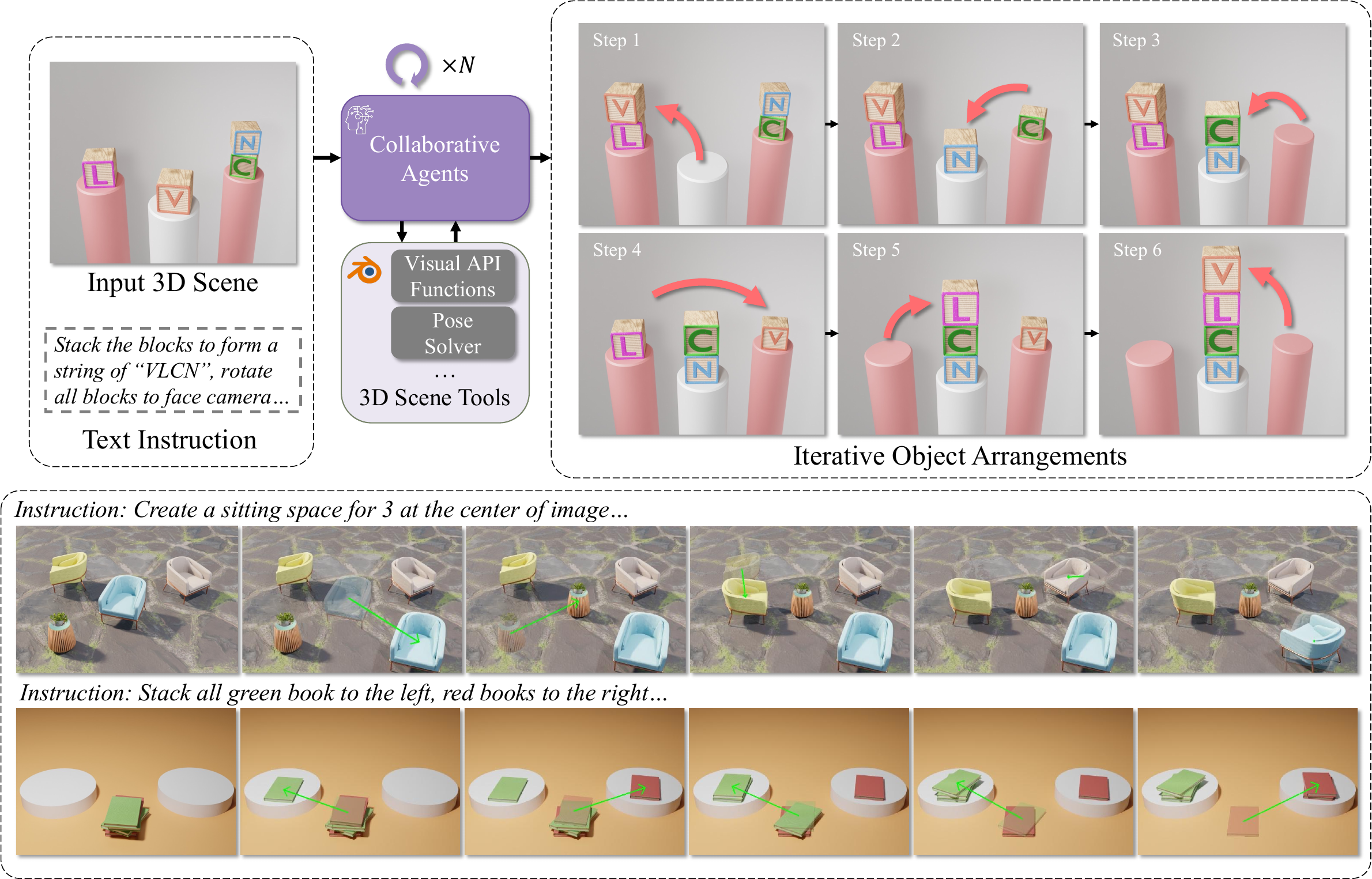}
\vspace*{-5pt}
\captionof{figure}{\method plans and executes multiple actions for complex object arrangement tasks given input image and user prompt.}
\vspace*{-5pt}
\end{strip} 

%% file: sec/0_abstract.tex
\begin{abstract}
Despite the remarkable progress of Multimodal Large Language Models (MLLMs) in 2D vision-language tasks, their application to complex 3D scene manipulation remains underexplored.
In this paper, we bridge this critical gap by tackling three key challenges in 3D object arrangement task using MLLMs.
First, to address the weak visual grounding of MLLMs, which struggle to link programmatic edits with precise 3D outcomes, we introduce an MCP-based API. This shifts the interaction from brittle raw code manipulation to more robust, function-level updates.
Second, we augment the MLLM's 3D scene understanding with a suite of specialized visual tools to analyze scene state, gather spatial information, and validate action outcomes.
This perceptual feedback loop is critical for closing the gap between language-based updates and precise 3D-aware manipulation.
Third, to manage the iterative, error-prone updates, we propose a collaborative multi-agent framework with designated roles for planning, execution, and verification. This decomposition allows the system to robustly handle multi-step instructions and recover from intermediate errors.
We demonstrate the effectiveness of our approach on a diverse set of 25 complex object arrangement tasks, where it significantly outperforms existing baselines.
\end{abstract}

%% file: sec/1_intro.tex
\section{Introduction}
\label{sec:intro}

Humans routinely rearrange objects in their environment, instinctively decomposing complex, multi-object tasks into sequential steps. For instance, moving a dining table naturally requires first clearing its surface and moving surrounding chairs. This fundamental ability to perform multi-step planning, grounded in deep spatial reasoning and a commonsense understanding of the physical world, is essential for executing complex arrangement tasks.

Recent Multimodal Large Language Models (MLLMs)~\cite{welsby2023chatgpt, team2023gemini, liu2023visual, bai2025qwen2} have enhanced human-like reasoning, enabling prior works~\cite{fireplace, 3dify, yang2024holodeck, layoutgpt, blenderalchemy, scanedit} to explore 3D object arrangement, which involves moving, rotating, or inserting objects to create a plausible layout. However, a key limitation of these approaches is their formulation of arrangement as a single-step process: starting from an initial scene, the agent performs one comprehensive edit to reach the goal state. The standard pipeline simply involves MLLM agents analyzing the 3D scene to generate this single edit.

In this work, we introduce a fundamentally more flexible paradigm. 
The core of our approach is the ability to elastically decompose tasks.
Our model is not confined to a single edit; instead it dynamically assesses the user's instruction and the scene's complexity.
For simple requests, it can generate a single-step solution, similar to the capabilities of prior work.
For complex tasks, it decomposes the instruction into a multi-step plan of sequential operations.
This ability to decide whether to perform a single edit or long-horizon sequence of actions—a capability that was by design infeasible for previous approaches—is what allows our model to mirror the human-level reasoning and planning.
Given a user instruction, whether abstract (e.g., ``Create a dining space") or detailed (e.g., ``Move the book to the shelf, then move the table"), the agent can now formulate a coherent, step-by-step plan and reliably execute each action in sequence to achieve the specified goal.

% In this work, we tackle one of the next-level challenges in 3D scene object arrangement: \textbf{Iterative 3D Object Arrangements}. In this problem, the model must handle complex arrangement tasks spanning multiple steps, where each step permits editing only \textbf{one object} at a time. This paradigm mirrors real-world human behavior, where we typically manipulate a single object at any given moment during long-horizon tasks. Given a user instruction—whether abstract (e.g., "Create a dining space") or detailed (e.g., "Move the book to the shelf, then move the table")—the agent must decompose the complex task into a detailed plan with step-by-step operations and reliably execute each arrangement action. This produces a sequence of coherent operations that ultimately achieves the specified goal.

% Compared to single-step formulations, this setting introduces three major challenges: 1) the iterative process demands high-quality arrangements not only in the final state but also at every intermediate state. Objects may be repositioned multiple times throughout execution, requiring plausible placement at each stage. 2) the multi-step nature creates interdependencies between actions—earlier decisions affect subsequent possibilities—making it difficult for agents to efficiently identify valid multi-step plans. 3) In cases where only abstract instructions are given (e.g. Create a dining space), the 

The fundamental requirement for this advanced task is a pipeline that operates with high-fidelity and robustness at every single step.
In iterative arrangement, unlike single-step tasks, errors in either scene analysis or execution can propagate and compound, quickly derailing the entire multi-step process.
Prior single-step approaches have employed various modalities for MLLM-based scene analysis, from raw visual representations~\cite{blenderalchemy, fireplace} to structured scene descriptions~\cite{scenelanguage, scenecraft}.
However, these methods strike a poor balance: they either overburden the MLLM with complex raw 3D data it cannot natively parse or provide simplified information that is insufficient for robust spatial reasoning.

This is the gap that recent MLLM tool calling \cite{yao2023react, pmlr-v235-kim24y, Liu2024ToolACEWT} and the development of the Model Context Protocol~\cite{mcp} have begun to address. 
These breakthroughs demonstrate a new way where MLLMs can be equipped to handle complex tasks~\cite{3dify, shen-etal-2025-zoomeye} that often lie beyond their core, encoded knowledge. This approach aids the MLLM by allowing to interact with external, use-defined application programming interfaces (APIs).
Inspired by these innovations, we address both the analysis and execution challenges. For interactive analysis, we equip our agents with a powerful visual API toolset. This allows them to actively query the environment, infer scene layouts from observations, and determine appropriate arrangement configurations without being overburdened. For reliable execution, we develop an advanced collision-free solver. This component ensures physical plausibility by translating the agent's high-level, symbolic intent into precise, valid numerical object poses.

%Furthermore, to ensure the physical plausibility of generated arrangements, we develop an advanced collision-free solver that computes numerical object poses from the agent's high-level intent.

The next critical challenge is scaling our tool-augmented capabilities to a multi-step scenario. This transition demands a sophisticated reasoning process: the agent must plan several moves ahead, ensuring each intermediate arrangement is not only physically plausible but also serves as a valid precondition for subsequent actions.
Simultaneously managing high-level, long-horizon strategy and low-level, step-by-step execution is a significant burden for a single agent.
We therefore separate these tasks by employing specialized agents with distinct responsibilities. 
One agent focuses on global planning across the entire task trajectory, while others handle the per-step execution and evaluation within the 3D scene.
This division is implemented in our collaborative multi-agent pipeline.

Another critical challenge is the combinatorial search space. 
Multi-step arrangement is inherently complex; the search space grows exponentially with each step, as every placement decision branches into numerous possible future states. This creates a tree of possibilities that quickly becomes intractable. 
To manage this complexity, we propose an effective search algorithm that strategically backtracks from unpromising paths, efficiently pruning the search tree and avoiding the trap of exhaustive enumeration.

We propose \method, a tailored agentic pipeline that leverages state-of-the-art MLLMs to generate accurate and plausible sequences of arrangements based on user instructions. We summarize our key contribution as: 

\begin{itemize}
\item\textbf{A tool-augmented MLLM} that integrates MCP-based visual tools with a constraint-based solver for efficient 3D scene analysis and physically-plausible arrangement;
\item\textbf{A collaborative multi-agent framework} that strategically separates responsibilities for global planning, step-by-step execution, and intermediate-state examination;
\item\textbf{An adaptive backtracking search algorithm} that efficiently navigates the exponential search space to find a valid path to the goal state.
\item\textbf{A new 3D object arrangment benchmark} designed to evaluate challenging 3D-aware object arrangements, on which our model significantly outperforms all the previous baselines.
\end{itemize}

%% file: sec/2_related_work.tex
\section{Related Works}

\paragraph{3D Object Manipulation and Arrangement}
The task of 3D object manipulation and arrangement is fundamental to both computer vision and robotics, as it enables agents to understand, interact with, and plausibly reorganize complex environments.
% While this area has been extensively studied, many image-based methods~\cite{visual_cot, zhang2023adding, bhat2024loosecontrol, nanobanana, liu2025step1x} focus on positional editing in 2D images. These approaches, however, lack robust 3D spatial context, incorporating either no 3D information or only limited data.
Conventional 3D object arrangement methods typically employ large 3D datasets~\cite{dai2017scannet, yang20253d, deitke2022, infinigen2023infinite, raistrick2024infinigen, li2023object, li2024controllable} to build data-driven models~\cite{2012-scenesynth, xu2013sketch2scene, fisher2015activity, wei2023lego, wang2018deep, wang2019planit, wang2021sceneformer}. However, these models are not guided by the common-sense reasoning of large language models, which limits their capability for general-purpose applications.
Following recent developments in MLLMs~\cite{welsby2023chatgpt, team2023gemini}, recent works~\cite{scenelanguage, scenecraft, yang2024holodeck, layoutgpt,  sun2025layoutvlm, wu2025human} have begun to leverage the reasoning capabilities of MLLMs to address object arrangement in a way that aligns with common sense. 
Since the standard MLLMs are not natively trained on 3D scene data at scale, these works enable the MLLM to access various intermediate representations for 3D scene analysis, such as textual scene description files~\cite{scenelanguage,de2024llmr}, code scripts~\cite{blenderalchemy, sun2025layoutvlm}, or rendered images~\cite{blenderalchemy,fireplace,scanedit}.
Among these, FirePlace~\cite{fireplace} and ScanEdit~\cite{scanedit} also apply dedicated solvers with geometric constraints to ensure physically plausible results.
However, all these works are limited to single-step operations. They lack the capability to handle complex multi-step scenarios where objects must be arranged sequentially with careful consideration of intermediate states. Our work directly addresses this gap by introducing an iterative framework that can decompose complex instructions into a coherent sequence of actions.
%a task that is by design challenging for previous single-step approaches.
\vspace{-2pt}

\paragraph{MLLMs for 3D}
The development of MLLMs extends large language models~\cite{team2023gemini, welsby2023chatgpt, touvron2023llama} to other modalities, including audio~\cite{ghosh2024gama}, image~\cite{liu2023visual, bai2025qwen2}, and video~\cite{videollm}.
Leveraging their powerful cross-modal reasoning capabilities, numerous works have developed systems to tackle various vision-language-based 3D tasks, such as geometry generation~\cite{nakayama2025aipparel,siddiqui2024meshgpt,lu2025ll3m,chen2025img2cad}, scene understanding~\cite{zhang2023video, Maaz2024VideoGPT+, huang2024chat, Inst3D-LMM, chen2024spatialvlm}, and appearance editing~\cite{3d-gpt,palandra2024gsedit}. Compared to methods that directly apply hard-coded algorithms~\cite{poole2022dreamfusion,instructnerf2023, kuang2023palettenerf, kuang2022neroic, liu2021editing, yuan2022nerf} or deep learning models on task-specific datasets~\cite{hong2023lrm,ren2024xcube,zhang2023dreamface,xu2023discoscene}, these MLLM-based approaches demonstrate better generalization and produce more semantically plausible results.
While benchmarks like BlenderGym~\cite{blendergym} were introduced to evaluate these 3D operations, both these existing works and their benchmarks were not designed for complex object arrangement. They provide limited support for the spatial reasoning required by an iterative, multi-object scenario. 
In contrast, our framework is uniquely designed for this long-horizon arrangement task, and to properly validate this new, complex capability, we also introduce a novel benchmark including 25 distinct object manipulation scenarios specifically designed to thoroughly test multi-step planning and reasoning capability.

%% file: sec/3_methods.tex
\begin{figure*}[t]
\centering
\includegraphics[width=\linewidth]{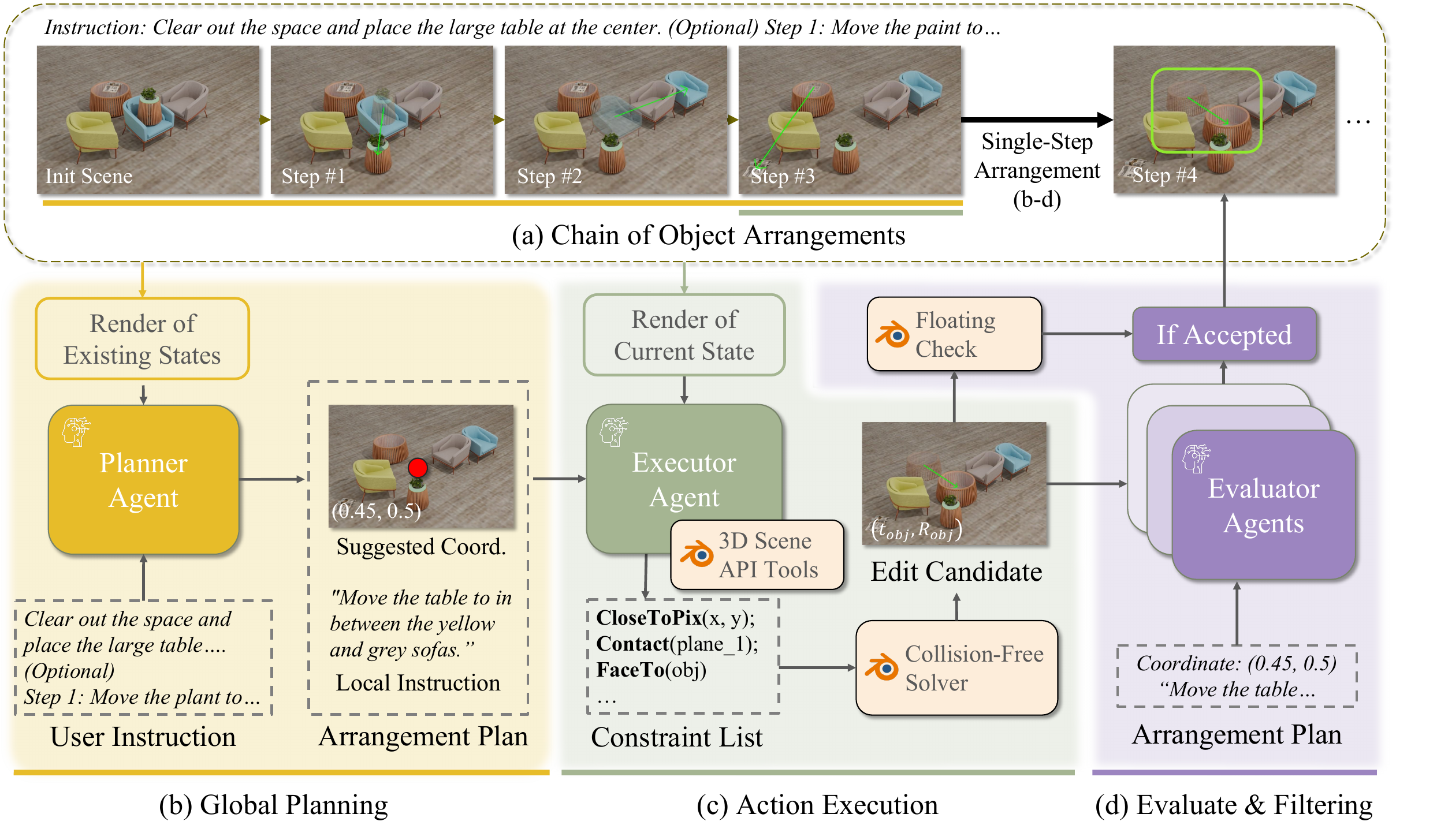}
\vspace{-10px}
\caption{
\textbf{\method Overview.} 
(a) Our method solves a long-horizon task through an iterative multi-agent process. 
(b) The \textit{Planner} agent examines the global context (the user instruction and all previous rendered states) to formulate a concrete plan for the current movement. 
(c) The \textit{Executor} implements this single-step plan in the 3D scene using API tools and solvers.
(d) A set of \textit{Evaluators} and an automatic floating check assess the execution quality. 
%If the step passes all checks, the new state is appended to the sequence history.
The entire ``Plan-Execute-Evaluate" loop repeats until the \textit{Planner} validates that the final arrangement fulfills the original user instruction.
}
\label{fig:multi_agent}
\vspace{-2mm}
\end{figure*}

\section{Problem Formulation}

We formulate our problem as follows: Given an image $I$ rendered from a fixed camera $C$ in an underlying 3D scene $S$, and a textual instruction $T$ (which may range from high-level goals to detailed per-step instructions), the model must output a sequence of single-object arrangements that accomplish the user's objective while maintaining physical plausibility throughout all intermediate states. 

We define three key objectives for these arrangements:

\begin{itemize}
\item \textbf{Collision-Free}: After each edit, the moved object should not collide with other objects in the scene.
\item \textbf{Floating-Free}: After each edit, every object in the scene (except for those originally floating) should be placed solidly on a surface.
\item \textbf{High Semantic Quality}: The arrangement should be semantically plausible and align well with the user's intent.
\end{itemize}

Unlike existing work that focuses on single-object insertion or offline multi-object placement, our problem addresses sequential, multi-step movements. At each step, the model must comprehend the relationships between all subtasks and objects to devise a globally coherent and actionable plan, then execute it reliably to prevent unacceptable cumulative errors. Additionally, the model must effectively search across different solution branches to identify valid paths to the goal, and when necessary, recover from dead ends caused by previous missteps.

To address these challenges, we introduce several key innovations detailed in the following sections. Section~\ref{sec:multi_agent} presents our collaborative MLLM agents for planning, placement, and evaluation; Section~\ref{sec:object_arrangement_framework} describes our object arrangement framework, including API-based visual tools and a constraint-based collision-free solver; and Section~\ref{sec:plan_searching} introduces our adaptive backtracking search algorithm for discovering valid multi-step plans.

\section{Multi-Agent Framework}
\label{sec:multi_agent}

Requiring a single MLLM to handle all 3D arrangement aspects introduces excessive processing burden and context length limitations. This problem is intensified in multi-step settings that demand global awareness. We therefore adopt a multi-agent approach with three specialized MLLMs~\cite{sun2025layoutvlm, 3d-gpt, fireplace}.
Our pipeline (Figure~\ref{fig:multi_agent}) consists of three agents per step: the \textit{\textbf{Planner}} extracts a concrete arrangement plan, the \textit{\textbf{Executor}} operates the 3D scene using our framework, and the \textit{\textbf{Evaluator}} validates the result.
To effectively distribute the workload, this architecture has two key designs:
First, the \textit{Planner} receives global context (i.e., information from other steps), while the \textit{Executor} and \textit{Evaluator} operate only within the local scope of the current step.
Second, only the \textit{Executor} interacts directly with the 3D scene; the other two agents observe rendered images.

\paragraph{Planning with Global Context}
The \textit{Planner} generates an executable action plan based on the user's instruction and a visual history, which consists of sequential renderings from previous steps. It analyzes this visual timeline to understand scene evolution and associates past states with the user's goal to determine the most appropriate next action.
Critically, the \textit{Planner} outputs both a textual instruction and an approximate target placement as a normalized 2D pixel coordinate
$\bm{c}=(x,y)$. 
We found this combination of language and spatial grounding to generate a less ambiguous plan.

\paragraph{Tool-Assisted Action Execution}
To address the challenges of direct MLLM execution, which demands simultaneous instruction grounding, 3D modeling, and physical plausibility checks, our \textit{Executor} defers these complexities to dedicated external modules.
Specifically, we equip the MLLM with a suite of MCP-based APIs for interactive tool usage.
The \textit{Executor} leverages these tools to project the \textit{Planner}'s instruction and 2D coordinate $c$ into the 3D environment.
The general execution pipeline proceeds as follows: 
First, it performs visual selection by querying the scene to localize a target object and receptacles.
Next, it translates intended spatial relationships into a set of geometric constraints~\cite{fireplace, scanedit}.
Finally, the \textit{Executor} invokes a optimization-based solver to compute an optimal, collision-free pose that guarantees physical validity.
Details of our visual selection, constraint formulation, and solver optimization are presented in Sec.~\ref{sec:object_arrangement_framework}.

\paragraph{Evaluation and Filtering}
The final agent, the \textit{Evaluator}, performs a visual check on the arrangement, examining its plausibility and consistency with the \textit{Planner}'s instruction.
Given the resulting image from the \textit{Executor}, the agent provides one of five categorical ratings: terrible, bad, fair, good, or excellent.
Due to the inherent uncertainty in current MLLMs, the \textit{Evaluator} can occasionally produce hallucinated evaluations (e.g., giving an ``excellent" rating to a completely incorrect placement). To mitigate this, we employ consensus-based filtering. This process involves polling multiple \textit{Evaluator} agents and calculating an average judgment score, mapped from -2 (terrible) to +2 (excellent). A solution is only accepted if this consensus score is positive.
Finally, in addition to the MLLM-based evaluation, we employ a rule-based check to detect physically implausible placements, such as floating objects. This deterministic check complements the agent's learned judgment by catching obvious physical errors that a model might overlook. Any movement that violates these physical constraints is automatically rejected.

%A rubric describing the standards for each category is provided to help the agent evaluate more accurately.

\begin{figure}[t]
  \centering
  \includegraphics[width=\linewidth]{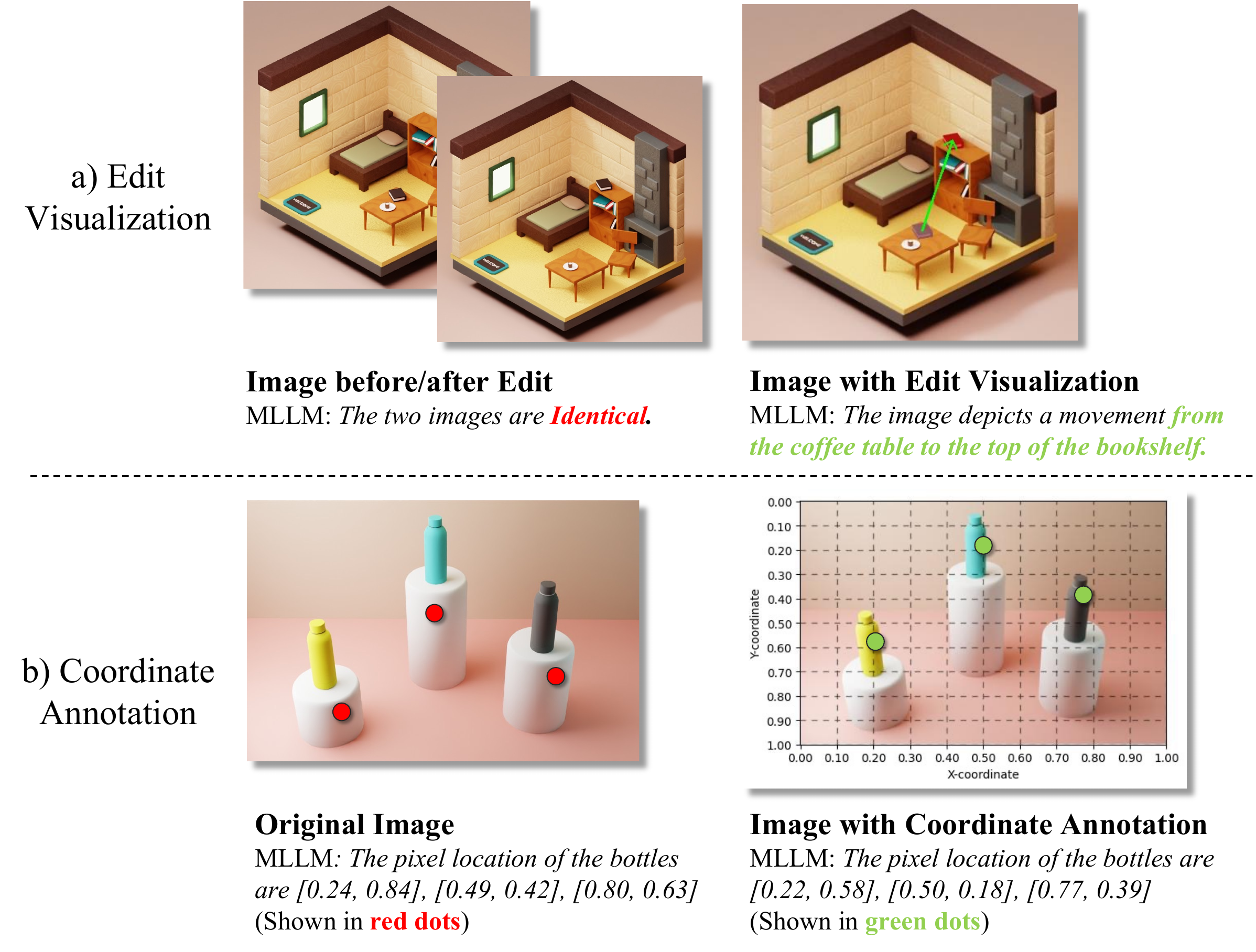}
  \vspace{-5px}
  \caption{
  \textbf{Impact of visual annotation.} 
  (a) Edit visualization improves the MLLM's spatial arrangement recognition; 
  (b) Coordinate annotations enhance its object localization accuracy.
  }
  \label{fig:enhanced_input}
  % \vspace{-15px} 
\end{figure}

\paragraph{Visual Annotation}
To enhance agent reasoning, we augment input images with two features (Fig.~\ref{fig:enhanced_input}). 
First, to provide precise spatial grounding, we add pixel coordinate labels and a dashed grid in the normalized pixel space.
This grid helps the \textit{Planner} to generate accurate target coordinates and for the \textit{Executor} to ground its tools.
Second, we visualize the completed action with an arrow connecting the object's original location to its target. This clearly displays the start and end states, providing context for the \textit{Planner} as it reviews the visual history and enabling the \textit{Evaluator} to effectively assess the most recent edit.

\paragraph{Adaptive Backtracking} \label{sec:plan_searching}
During multi-step execution, errors from previous steps can render subsequent steps infeasible, leading to dead ends. For example, poor placement of initial objects on a table may leave insufficient space for later ones. To overcome this, we introduce an adaptive backtracking search algorithm, inspired by classic methods~\cite{knuth1977fast, chiu1989analysis}, to efficiently recover from these errors. Our algorithm maintains an \textit{anchor} step, which serves as the restart position for failed attempts. This anchor is adaptively reset: it moves to half the current depth when an attempt fails, or it advances when the action sequence reaches a new maximum length from the whole run (e.g. successfully placed three bottles on the table, while previous attempts placed at most two).

% While our collaborative agents and object arrangement framework can efficiently execute individual steps, applying them to multi-step tasks presents a critical challenge: errors from previous steps can render subsequent steps infeasible. In such scenarios, the naive approach of repeatedly executing the pipeline until the goal is reached fails. For example, when placing multiple objects on a small table, poor placement of initial objects may leave insufficient space for the remaining ones. To address this, we draw inspiration from classic search algorithms~\cite{knuth1977fast, chiu1989analysis} and develop an efficient search algorithm with adaptive backtracking that can recover from dead ends caused by previous errors.

% Specifically, we maintain an \textit{anchor} step that serves as the restart position for failed attempts. The anchor is reset whenever the current attempt fails (by halving the current depth) or when the arrangement sequence reaches a new maximum length. This design enables adaptive exploration of previous steps, prioritizing closer positions before backtracking further. Our adaptive backtracking requires significantly less search attempts compared to restarting from the initial state after each failure.

% To achieve robust object arrangement with minimal artifacts, we adopt a constraint-based optimization approach following recent state-of-the-art methods~\cite{fireplace, scanedit}. Rather than directly predicting numerical pose values, the \textit{Executor} outputs a set of symbolic constraints that are subsequently optimized to determine the target object pose. This formulation significantly reduces placement artifacts while providing an interpretable interface for MLLM-based reasoning. Our framework supports the following constraint types:

\section{Multi-Tool Library}
\label{sec:object_arrangement_framework}
As described in Sec.~\ref{sec:multi_agent}, the \textit{Executor} agent's primary role is to translate the \textit{Planner}'s 2D-based instruction into a physically valid 3D action. To bridge this 2D-to-3D gap, the \textit{Executor} is equipped with a visual tool library that defines a clear, step-by-step workflow:

\begin{enumerate}
    \item {Visual Probing:} The agent queries the 3D scene to ``see" and identify the 3D objects and planes.

    \item {Constraint Formulation:} The agent assembles a placement plan as a list of geometric constraints.

    \item {Optimization:} The constraints are passed to a solver, which computes the optimal valid 3D pose.
\end{enumerate}
\vspace{-3mm}
% This section details these three components.

\begin{figure}[t]
  \centering
  \includegraphics[width=\linewidth]{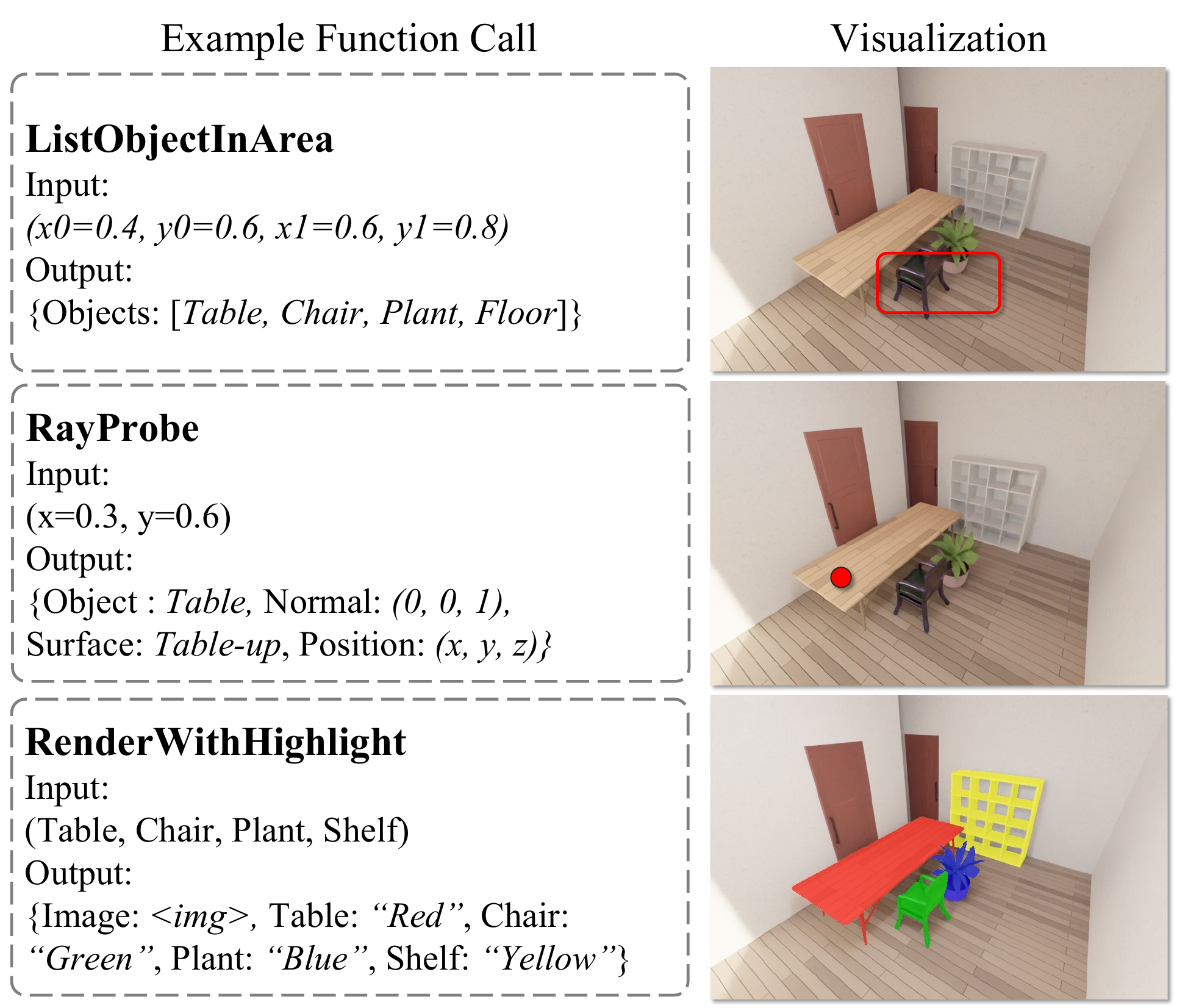}
  \vspace{-10px}
  \caption{\textbf{Visual Probing Tools.} For each API, we present an example function call format, along with a result visualization.
  }
  \label{fig:visual_func}
  % \vspace{-12px}
\end{figure}

\paragraph{Visual Probing}
We equip the agent with three visual probing tools, to interactively analyze the 3D scene layout.

\begin{itemize} 
\item \textsc{\textbf{ListObjectsInArea}}: Returns object names within or partially overlapping a specified image area. This helps the MLLM identify objects visible in that region.
\item \textsc{\textbf{RayProbe}}: Casts a ray from the camera at a given pixel coordinate and returns information about the first hit point. This includes the 3D location, the name of the hit object, and the name and normal of the hit plane.
\item \textsc{\textbf{RenderWithHighlight}}: Renders an instance segmentation image with specified objects highlighted. This function helps the MLLM confirm the correlation between object names and their visual representations.
\end{itemize}

During a typical \textit{Executor} run, we observe that the agent calls \textsc{\textbf{ListObjectsInArea}} and \textsc{\textbf{RayProbe}} to retrieve the names of relevant objects and planes, including the object to be edited and any reference objects. 
If the name alone is ambiguous (e.g., multiple objects named ``Book"), the agent will then call \textsc{\textbf{RenderWithHighlight}} to visually discern the correct target object from the others.
\vspace{-3mm}

\paragraph{Constraint Formulation}
After identifying the necessary 3D elements, the agent assembles a placement plan as a list of constraints. The supported constraints are:

\begin{itemize}
\item $\textsc{\textbf{CloseToPix}}(obj, x, y)$: The placed object $obj$ should be located close to the pixel coordinate $(x, y)$ in camera $C$'s view.
\item $\textsc{\textbf{Contact}}(obj, dir, p)$: The plane of the object $obj$'s bounding box facing $dir$ should contact the plane $p$.
\item $\textsc{\textbf{NoOverhang}}(obj, dir, p)$: The plane of the object's bounding box facing $dir$ should be fully inside the convex hull of plane $p$.
\item $\textsc{\textbf{Distance}}(obj, obj_2, dist)$: The Euclidean distance between $obj$ and $obj_2$ should be close to $dist$.
\item $\textsc{\textbf{FaceTo}}(obj, obj_2/p)$: The facing direction of object $obj$ should point towards the center of $obj_2$, or align with the normal direction of plane $p$.
\item $\textsc{\textbf{Rotate}}(obj, degree)$: The object should rotate counterclockwise by $degree$.
\end{itemize}

Together, these constraints form a comprehensive vocabulary that ensures strong geometric plausibility and is applicable to a wide variety of 3D object arrangement tasks.

\paragraph{Optimization}
\label{sec:collision_free_solver}
Given that the target pixel $c$ and constraints from previous steps already define a small search space,  we adopt an efficient, sampling-based approach. Inspired by preconditioning techniques in numerical analysis~\cite{Axelsson_1994,Chen_2005}, we first randomly perturb the target pixel coordinate $c$ into a batch of variants. Next, we apply the solver to each variant, generating a set of candidate poses distributed around the original target. The solver employs an AdamW~\cite{Loshchilov2017DecoupledWD} optimizer to minimize constraint losses. Following optimization, we validate each solved candidate by recalculating its constraint losses in the original pixel coordinate space, and remove any candidates whose error exceeds the threshold $\tau$.
Finally, a collision detector filters these candidates, and we select the valid, collision-free pose with the lowest constraint error. 
This method, detailed in Alg.~\ref{alg:solver}, efficiently ensures a non-colliding solution.

% Once the \textit{Executor} completes its execution, we parse the constraint list from its output, treating each constraint as a differentiable loss function. While gradient-based optimization could directly minimize these losses, this straightforward approach frequently produces poses that collide with other objects. Alternatively, integrating collision avoidance as an additional differentiable loss term would require gradient computation at each iteration, which is a computationally expensive operation in complex scenes.

% Given a target pixel coordinate $c$ and constraints composed by previous steps, the search space for valid object poses is small. Therefore, we opt for an intermediate solution. Inspired by preconditioning techniques in numerical analysis~\cite{Axelsson_1994,Chen_2005}, we first randomly perturb the target pixel coordinate in the \textsc{CloseToPix} constraint into a batch of variants. We then apply the solver to each variant. This creates a batch of valid poses distributed around the original target. Finally, we apply a collision detector to these poses and select the valid one that is closest to the target pixel. 
% Alg.~\ref{alg:solver} shows the pseudocode of our solver. This approach ensures the non-collision property of the solution while maintaining the efficiency of the method.

\RestyleAlgo{ruled}
\SetKwInput{KwInput}{Input}
\SetKwInput{KwParameter}{Parameter}
\SetKwInput{KwOutput}{Output}
\SetKwComment{Comment}{/* }{ */}

\begin{algorithm}[hbt!]
\caption{Constraint-based Pose Solver}\label{alg:solver}
\KwParameter{pixel coordinate variance $\sigma_{pix}$, loss threshold $\tau$}
\KwInput{Suggested pixel coordinate $\bm{c}$, Constraint list $D=\{d_{1,...,l}\}$}
\KwOutput{The target object pose $T_{obj}=(t_{obj},R_{obj})$}

$\bm{c}_{1,...,n}'\gets \mathrm{Sample}(\mathcal{N}(\bm{c}, \sigma_{pix}))$\;

$T_{1,...,n} \gets \mathrm{Optimize}(D, \bm{c}'_{1,...,n})$ \;
\Comment{Pre-filtering with constraint errors}
$T^{pre}_{1,...,m} \gets \mathrm{FilterByError}(T_{1,...,n}, D, \bm{c}, \tau)$\;

\For{$T_i$ in $\mathrm{SortByError}(T^{pre}_{1,...,m})$}
{
    \uIf{not $\mathrm{CollisionDetected}(T_i)$}
    {
        \Return $T_i$\;
    }
}
\Return ``Failed"\;
% {
%     \Comment{Generate chunk prediction} 
%     $\bm{z}^{ch} \gets \bm{z}_{ch,...,ch+C-1}$\;
%     $\bm{\mathcal{G}}^{ch} \gets \mathcal{D}(\bm{z}^{ch})$\; 
    
%     \Comment{Loss on decoded frames} 
%     $l \gets \mathcal{L}_{pix}(\bm{\mathcal{G}}^{ch})$\;
%     $\bm{g}^{ch} \gets \texttt{Autograd}(l, \bm{z}^{ch})$\;
%     \Comment{SG means stop gradient} 
%     $\mathcal{L}_{def} \gets \mathcal{L}_{def} + \frac{1}{K}\texttt{Sum}(\texttt{SG}(\bm{g}^{ch}) \cdot \bm{z}^{ch})$ \; 
% }
% \Return $\mathcal{L}_{def}$

% $\mathcal{L}_{def} \gets 0$\;
% $\bm{z}_{1,...,K}\gets f_{\theta}(\bm{I}_{1,...,K})$\;
% \For{$ch$ in \texttt{Range}(\texttt{start}=1, \texttt{end}=K, \texttt{step}=C)}
% {
%     \Comment{Generate chunk prediction} 
%     $\bm{z}^{ch} \gets \bm{z}_{ch,...,ch+C-1}$\;
%     $\bm{\mathcal{G}}^{ch} \gets \mathcal{D}(\bm{z}^{ch})$\; 
    
%     \Comment{Loss on decoded frames} 
%     $l \gets \mathcal{L}_{pix}(\bm{\mathcal{G}}^{ch})$\;
%     $\bm{g}^{ch} \gets \texttt{Autograd}(l, \bm{z}^{ch})$\;
%     \Comment{SG means stop gradient} 
%     $\mathcal{L}_{def} \gets \mathcal{L}_{def} + \frac{1}{K}\texttt{Sum}(\texttt{SG}(\bm{g}^{ch}) \cdot \bm{z}^{ch})$ \; 
% }
% \Return $\mathcal{L}_{def}$
\end{algorithm}
% Inspired by classic search algorithms in string processing and computer networks\cite{knuth1977fast, chiu1989analysis}, we developed an efficient search algorithm that is robust to invalid branches, based on an adaptive backtracking technique. As Alg.\ref{alg:plan_search} shows, we maintain an \textit{anchor} step during the search, which serves as the restart position for any failed attempt. This anchor is reset whenever the current attempt fails (e.g., by halving the current depth) or when the arrangement chain successfully reaches a new maximum length.

% This design allows the system to adaptively check different previous steps, starting from the closer ones to the further ones. This adaptive backtracking technique significantly reduces the number of attempts compared to simply restarting from the initial state when an attempt fails, as shown in our experiments. \zhengfei{get the number in our experiment}

%% file: sec/4_experiments.tex
\begin{figure}[t]
  \centering
  \includegraphics[width=\linewidth]{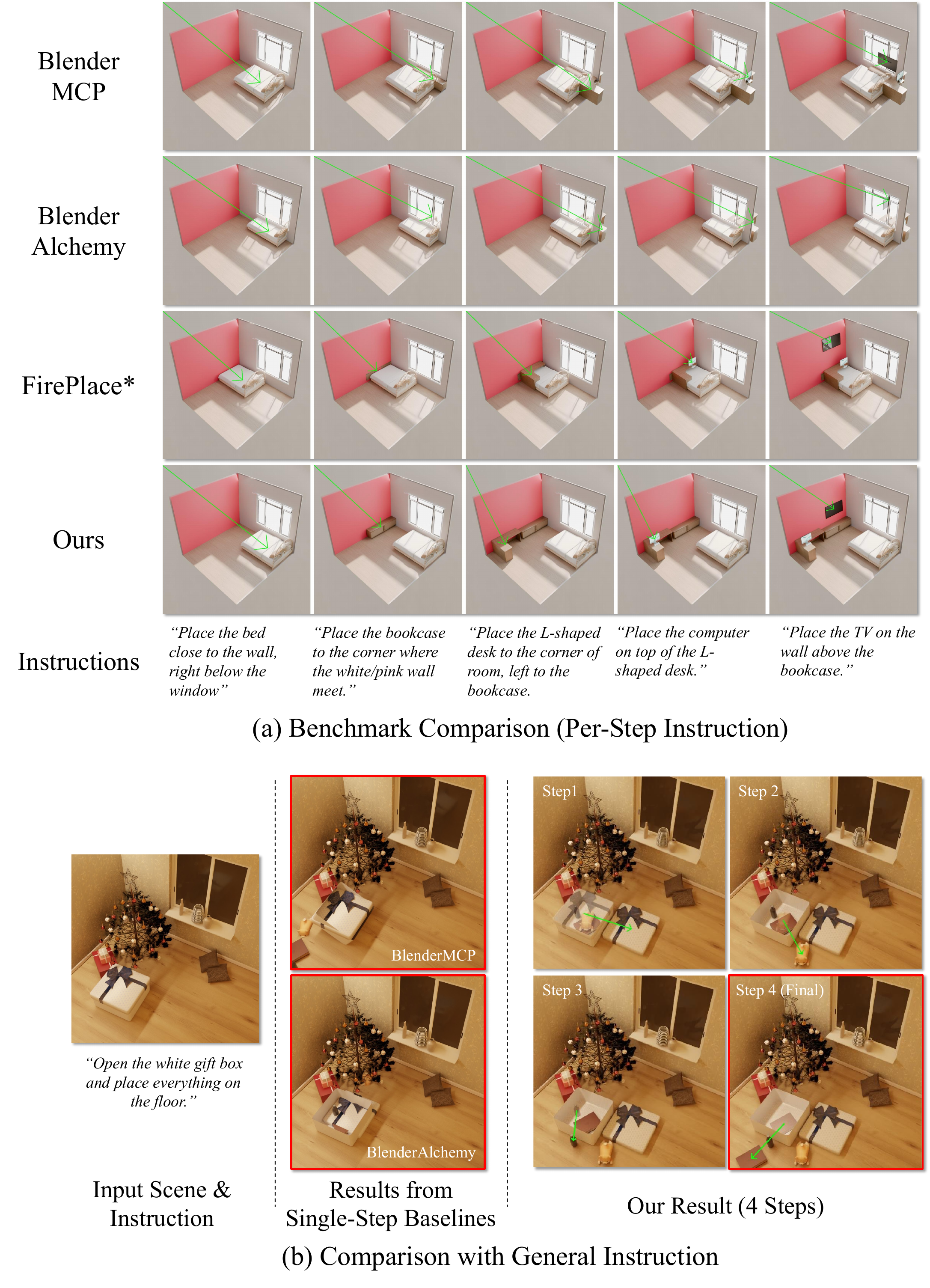}

  \caption{
  \textbf{Qualitative Comparisons.} 
  We compare results using two types of guidance: detailed per-step instructions (\textit{Top}) and general, abstract instructions (\textit{Bottom}).
  \textit{Top}: When given per-step instructions, our method outperforms the baselines. \textit{Bottom}: Using general instructions, our multi-step approach produces reasonable intermediate states and yields a significantly better final state (marked in red) compared to single-step baselines. 
  }
  \label{fig:qualitative_comparison}
  \vspace{-1px}
\end{figure} 
\section{Experiments}

\begin{figure*}[t]
  \centering
  \includegraphics[width=\linewidth]{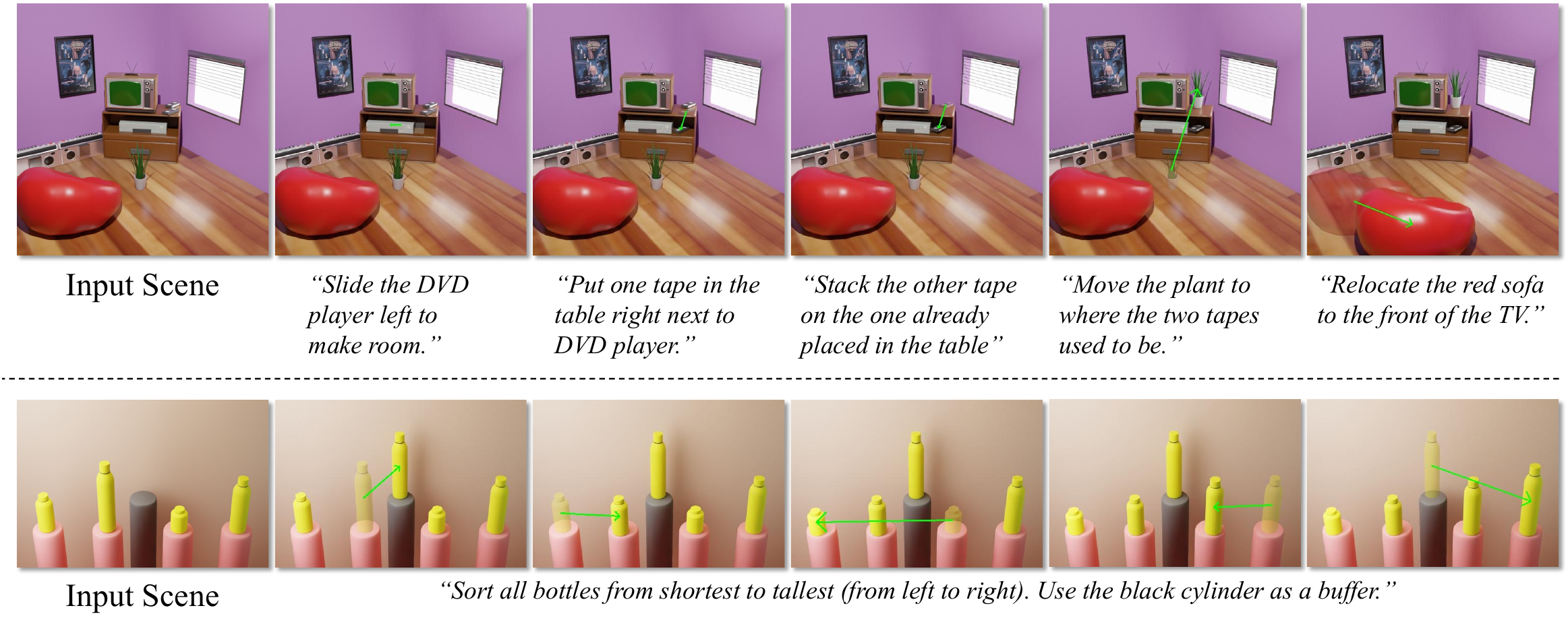}
  \vspace{-10px}
  \caption{\textbf{More Visual Results.} Our model tackles complex arrangement tasks with per-step prompts (\textit{top}) or general instructions (\textit{bottom}). }
  \label{fig:qualitative}
  \vspace{-10px}
\end{figure*}

We conduct all experiments using Gemini-2.5-pro, a state-of-the-art MLLM. 
Our object arrangement framework, including the constraint-based solver and API visual functions, is implemented using Blender's Python extension packages integrated with the Blender-MCP library~\cite{blendermcp}. 
For each arrangement step, we employ 3 \textit{Evaluator} agents and execute 4 parallel attempts, selecting the valid solution with the highest average evaluation score. If no solution is found in all attempts, the step is marked as failed and the system proceeds with backtracking.

\subsection{Dataset and Metrics}

We evaluate our model's ability to produce plausible and precise multi-step arrangements through comprehensive benchmark comparisons against baseline approaches. Our evaluation dataset comprises 25 carefully curated scenes sourced from BlenderKit~\cite{blenderkit}, InfiniGen~\cite{infinigen2023infinite, raistrick2024infinigen}, and BlenderGym~\cite{blendergym}, encompassing 111 unit tasks in total. To ensure fair comparison, all baseline methods receive identical reference per-step instructions and execute the same sequence of operations. The task design deliberately introduces interdependencies where subsequent steps can be affected by preceding operations without proper execution.

We assess performance at every step using the following metrics, averaged across all tasks:

\begin{itemize}
\item \textbf{Collision Rate}: The proportion of steps that result in a collision between the moved object and other elements.
\item \textbf{Floating Rate}: The proportion of steps that cause any object in the scene to become unsupported (i.e., ``floating"). 
%Note that an arrangement can affect all elements within the scene.
\item \textbf{Plausibility}: The physical realism of the arrangement, assessed by an MLLM, scored from 0 (poor) to 4 (great).
\item \textbf{Consistency}:The alignment between the user's instructions and the resulting arrangement, assessed by an MLLM and scored from 0 to 4.
\end{itemize}

\input{tables/benchmark}

\subsection{Baselines Comparisons}

We compare our \method against three recent MLLM-based approaches, selected to represent different components of our system:
\textbf{BlenderAlchemy}~\cite{blenderalchemy} employs MLLM agents in an executor-evaluator loop to edit Blender scenes via Python scripts. This aligns with parts of our multi-agent structure but lacks the interactive visual tools and constraint-based solver used by our Executor agent. We provide object poses as input to its script representation.
\textbf{Fireplace}~\cite{fireplace} represents the state-of-the-art in using a constraint-based solver with MLLMs for object placement, which is analogous to our Executor's core solving capability. However, it operates as a standalone solver and lacks our system's broader multi-agent framework (i.e., the Planner and Evaluator agents) for global context and multi-step reasoning. We reimplemented the method for our setting.
\textbf{Blender-MCP}~\cite{blendermcp} enables MLLM interaction with 3D scenes through basic callable functions (\textsc{GetSceneInfo}, \textsc{GetObjectInfo}, \textsc{RenderImage}, and \textsc{ExecuteCode}), representing a simpler, tool-only baseline. It lacks our framework's specialized agent roles (Planner, Evaluator), constraint-based solver, and multi-step reasoning capabilities. We configure this baseline to use its default function set.

Table~\ref{tab:quantitative_comparison} presents the quantitative results, demonstrating that our approach substantially outperforms all baselines. Our method not only generates more realistic and accurate movements but also precisely eliminates collisions and floating artifacts, thanks to our collision-free solver and filter techniques.
Figure~\ref{fig:qualitative_comparison} shows qualitative comparisons on benchmark and general instruction examples. In the first example, BlenderAlchemy and Blender-MCP fail completely without effective API tools and solvers. Fireplace successfully moves the bed but subsequent placements collide with it. Our method achieves artifact-free results. In the second example, single-step baselines fail on complex tasks, while our multi-step approach produces high-quality results.

% \subsection{Human Evaluation}
% Beyond automated MLLM assessment, we conduct a user study with \zhengfei{TODO: add number} participants comparing our model against baselines. Participants view outputs from our model alongside baseline results and judge which demonstrates better physical correctness, visual plausibility, and consistent with the instructions. We record the preference rate on our model over each baseline. As shown in Tab.~\ref{tab:quantitative_comparison}, participants overwhelmingly prefer our model's outputs by a substantial margin.
% % \input{tables/human_study}

\subsection{Human Evaluation}
In addition to MLLM evaluations, we also conduct a human study involving 30 subjects for evaluating our models with baselines. Specifically, we provide the result of our model and result of the baselines, and let the subjects to decide which one looks more physically correct, visually plausible and consistent to the instructions. We collect the winning rate of our model competing with each baselines. As Tab.~\ref{tab:human_study} shows, users generally prefer our models's result than the others with a huge margin.

\input{tables/human_study}

\subsection{Ablation Study}

We conduct ablation experiments on the same benchmark to validate the contribution of each architectural component. We compare our full model against five variants:
\textbf{Model w/o Multi Tool Library} removes the constraint-based solver and visual API, using Blender-MCP~\cite{blendermcp} functions instead.
\textbf{Model w/o Backtracking} eliminates adaptive backtracking and always proceeds to the next step, selecting the best solution even with artifacts (solver allows collisions).
\textbf{Model w/o MCP Tools} variant removes all tools requiring MCP~\cite{mcp} (including the original Blender-MCP~\cite{blendermcp} functions). Instead, it requires the model to directly predict raw Blender~\cite{blender} Python scripts for object arrangement.
\textbf{Model w/o Planner Coordinates} variant removes coordinate guidance from the \textit{Planner} agent.
\textbf{Model w/ Single Agent} replaces our collaborative multi-agent system with a single MLLM agent.

Table~\ref{tab:ablation_study} demonstrates that the complete model achieves optimal performance across all metrics. Removing the solver and visual tools substantially degrades plausibility and quality, though performance still exceeds the original Blender-MCP baseline due to our multi-agent design and backtracking mechanism. Without backtracking, the model occasionally accepts suboptimal solutions, resulting in small but non-zero collision and floating rates—though the overall pipeline robustness keeps these occurrences rare in the benchmark. The single-agent variant maintains zero collision and floating artifacts but exhibits reduced visual quality and textual consistency, confirming the importance of our multi-agent decomposition for preserving reasoning quality in complex multi-step scenarios.

\input{tables/ablation_study}

\subsection{Visual Results}
\begin{figure}[t]
  \centering
  \includegraphics[width=\linewidth]{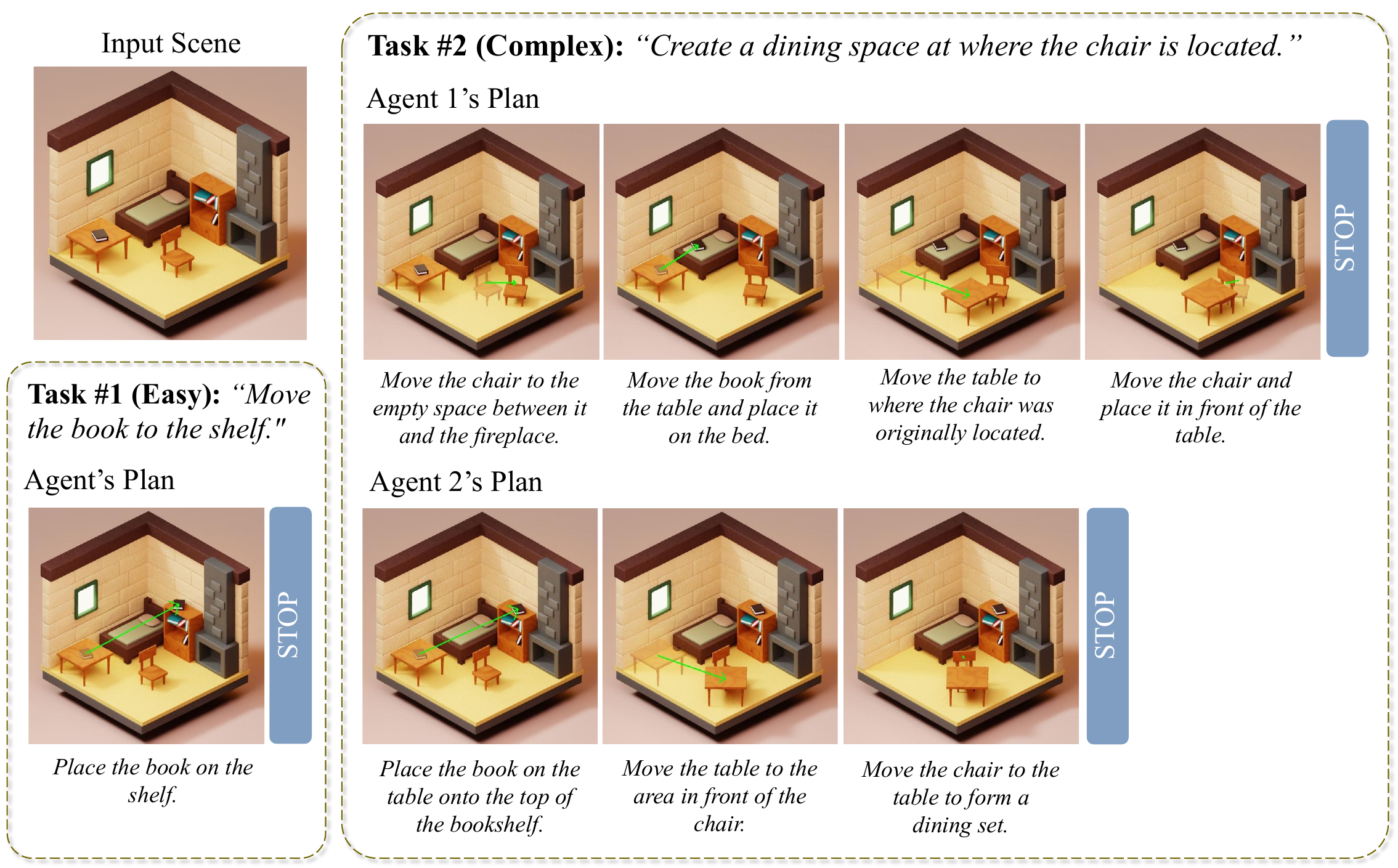}
  \vspace{-5px}
  \caption{\textbf{Adaptive Planning.} Our model plans adaptively for different tasks and can generate diverse plans for the same objective. The plan output is shown below the corresponding edited image.}
  \label{fig:diverse_plan}
  \vspace{-5px}
\end{figure}

We present additional qualitative results demonstrating our method's performance across various tasks. Figure~\ref{fig:qualitative} illustrates our model's ability to handle complex scenarios requiring robust placement pipelines and effective global reasoning across multiple steps. Figure~\ref{fig:diverse_plan} further demonstrates the model's flexibility in generating diverse plans: It adapts its planning to different tasks and produces diverse solutions for identical objectives across runs.

%% file: tables/benchmark.tex
\begin{table}[t]
\centering
\setlength{\tabcolsep}{2.5pt}
\renewcommand{\arraystretch}{1.4}
\small
\begin{tabular}{lcccc}
\toprule
% Method & Time & Coll.\%$\downarrow$ & Float.\%$\downarrow$ & Plausibility $\uparrow$ & Consistency$\uparrow$ & Physical Correctness & Plausibility & Consistency \\
Method & Coll.\%$\downarrow$ & Fl.\%$\downarrow$ & Plausibility $\uparrow$ &
Consistency$\uparrow$  \\
\midrule
Blender-MCP &  
% &  &  &  &  \\
0.459 & 0.774 & 3.348 & 2.973 \\
BlenderAlchemy & 
0.631 & 0.676 & 3.368 & 2.770 \\
FirePlace* &  
0.513 & 0.225 & 3.515 & 3.135 \\
\midrule
Ours &  
\textbf{0.000} & \textbf{0.000} & \textbf{3.796} & \textbf{3.592} \\
\bottomrule
\end{tabular}
\caption{\textbf{Quantitative Comparisons.}
We compare our method against baselines using several metrics:
the average rate of collision (\textit{Coll.\%}), the rate of floating (\textit{Fl.\%}), va isual plausibility score (\textit{Plausibility}), and a consistency score between the text instruction and edited image (\textit{Consistency}).
% The latter two metrics are evaluated using an MLLM.
}
\label{tab:quantitative_comparison}
\end{table}

% \begin{table*}[ht]
% \centering
% \setlength{\tabcolsep}{2.5pt}
% \renewcommand{\arraystretch}{1.4}
% \small
% \begin{tabular}{l c ccccc c ccc}
% \toprule
% % Method & Time & Coll.\%$\downarrow$ & Float.\%$\downarrow$ & Plausibility $\uparrow$ & Consistency$\uparrow$ & Physical Correctness & Plausibility & Consistency \\
% \multirow{2}{*}{Method} && \multicolumn{5}{c}{Quantitative Benchmark}  && \multicolumn{3}{c}{Human Evaluation (Ours vs. Baseline)} \\
% \cline{3-7} \cline{9-11} 
% &&& Coll.\%$\downarrow$ & Float.\%$\downarrow$ & Plausibility $\uparrow$ &
% Consistency$\uparrow$ &&
% Physical Correctness & Plausibility & Consistency \\
% \midrule
% Blender-MCP &&  
% % &  &  &  &  \\
% & 0.459 & 0.774 & 3.348 & 2.973 \\
% BlenderAlchemy && 
% & 0.631 & 0.676 & 3.368 & 2.770 \\
% FirePlace* &&  
% & 0.513 & 0.225 & 3.515 & 3.135 \\
% \midrule
% Ours &&  
% & \textbf{0.000} & \textbf{0.000} & \textbf{3.796} & \textbf{3.592} &&
% N/A & N/A & N/A \\
% \bottomrule
% \end{tabular}
% \caption{\textbf{Comparison results on our quantitative benchmark.}
% We compare our method against BlenderAlchemy and BlenderMCP on the average rate of collision (\textit{Call. \%}), rate of floating (\textit{Float. \%}), visual plausibility score (\textit{Plausibility}) and consistency score between edited image and text instruction  (\textit{Consistency}) from MLLM.
% Our approach outperforms all baselines and variants. *We re-implemented FirePlace adapted to our task since its code is not released.}
% \label{tab:quantitative_comparison}
% \end{table*}

%% file: tables/human_study.tex
\begin{table}[ht]
\centering
\setlength{\tabcolsep}{2.5pt}
\renewcommand{\arraystretch}{1.4}
\small
\begin{tabular}{lccc}
\toprule
\multirow{2}{*}{Method} & \multicolumn{3}{c}{Win \% / Tie \%} \\
& Const. & Plaus. & Phys. \\
\midrule
Ours vs. BlenderAlchemy~\cite{blenderalchemy} & 
\textbf{62.1}/22.7 & \textbf{65.9}/24.7 & \textbf{70.0}/20.0   \\
Ours vs. Blender-MCP~\cite{blendermcp} & 
\textbf{60.5}/25.0 & \textbf{59.0}/25.8 & \textbf{62.9}/26.1   \\
Ours vs. FirePlace*~\cite{fireplace} & 
\textbf{58.8}/30.2 & \textbf{54.4}/35.0 & \textbf{58.0}/31.5   \\
\bottomrule
\end{tabular}
\caption{\textbf{Results on Human Study.}
We conduct a human study to compare our model with baselines on three domains: Instruction Following (\textit{Const.}), Visual Plausibility (\textit{Plaus.}) and Physical Plausibility (\textit{Phys.}). Our model is significantly more favorable to the users than the baselines in all aspects. We re-implemented Fireplace with adaptation to our task since its code is not released.}
\label{tab:human_study}
\vspace{-10pt}
\end{table}

%% file: tables/ablation_study.tex
\begin{table}[t]
\centering
\setlength{\tabcolsep}{2.5pt}
\renewcommand{\arraystretch}{1.4}
\small
\begin{tabular}{lcccc}
\toprule
Method & Coll.\%$\downarrow$ & Fl.\%$\downarrow$ &  Plaus.$\uparrow$ & Const.$\uparrow$ \\
\midrule
w/o Multi-Tool Library & 0.495 & 0.711 & 3.484 & 3.103 \\
w/o Backtracking & 0.036 & 0.054 & 3.703 & 3.549 \\
w/o MCP Tools  & 0.603 & 0.738 & 3.357 & 3.067 \\
w/o Planner's Coordinates & \textbf{0.000} & \textbf{0.000} & 3.772 & 3.526 \\
Single Agent & \textbf{0.000} & \textbf{0.000} & 3.623 & 3.328 \\
\midrule
Ours & \textbf{0.000} & \textbf{0.000} & \textbf{3.796} & \textbf{3.592} \\
\bottomrule
\end{tabular}
\caption{\textbf{Ablation Study Results.} Our complete model achieves substantially better performance than any ablated variant.}
\label{tab:ablation_study}
\vspace{-10pt}
\end{table}

% \begin{table}[ht]
% \centering
% \setlength{\tabcolsep}{2.5pt}
% \renewcommand{\arraystretch}{1.4}
% \small
% \begin{tabular}{lcccc}
% \toprule
% Method & Coll.\%$\downarrow$ & Float.\%$\downarrow$ &  Plaus.$\uparrow$ & Const.$\uparrow$ \\
% \midrule
% Placer-only & - & - & - & - \\
% Placer + Planner  & - & - & - & - \\
% Placer + Evaluator  & - & - & - & - \\
% \midrule
% w/o Toolbox & - & - & - & - \\
% Visual probes Only & - & - & - & - \\
% \midrule
% w/o Backtracking & - & - & - & - \\
% \midrule
% Ours & 0.000 & 0.000 & 3.796 & 3.592 \\
% \bottomrule
% \end{tabular}
% \caption{\textbf{Ablation Study.}}
% \label{tab:ablation_study}
% \end{table}

% \begin{table}[ht]
% \centering
% \setlength{\tabcolsep}{2.5pt}
% \renewcommand{\arraystretch}{1.4}
% \small
% \begin{tabular}{lcccc}
% \toprule
% Method & Coll.\%$\downarrow$ & Float.\%$\downarrow$ &  Plaus.$\uparrow$ & Const.$\uparrow$ \\
% \midrule
% w/o Visual Probe Tools & - & - & - & - \\
% w/o Constraints-based Solvers & - & - & - & - \\
% \midrule
% Single Agnet & - & - & - & - \\
% w/o Evaluators & - & - & - & - \\
% w/o Planner & - & - & - & - \\
% Ours & 0.000 & 0.000 & 3.957 & 3.914 \\
% \bottomrule
% \end{tabular}
% \caption{\textbf{Ablation Study.}}
% \label{tab:ablation_study}
% \end{table}

%% file: sec/5_discussion.tex
\section{Conclusion}
 We propose \method, a robust iterative object arrangement pipeline that addresses multi-step 3D arrangement tasks from textual input. Equipped with novel visual API tools, a collision-free solver, collaborative agents, and adaptive backtracking, our model effectively handles complex tasks and substantially outperforms state-of-the-art baselines. Meanwhile, to maintain optimal MLLM reasoning performance with reasonable context length, our current system is designed to support single-camera views only. This limits performance on tasks requiring novel view observations, for example, placing a chair to the unseen space behind a wall. We believe our pipeline can be extended to multi-view scenarios as long-context MLLMs continue to develop.

 \subsection{Acknowledgement}
 This work was supported by computational resources provided by Google Cloud. We thank Alireza Fathi, Yanan Bao, and Ian Huang for their assistance in reproducing the results from Fireplace, as well as Jiacheng Chen for his technical support with Blender scripting.

%% file: sec/X_suppl.tex
\clearpage
\setcounter{page}{1}
\maketitlesupplementary

\section{Comparison on Abstract Prompts}

We demonstrate that our proposal can flexibly handle both abstract prompts (requiring model to decompose the task into action plans) and detailed multi-step instructions (requiring the model to faithfully execute instructions sequentially).
Qualitative examples for both cases are provided in Fig. \ref{fig:supp_qual_comp_per_step_1}, Fig. \ref{fig:supp_qual_comp_per_step_2}, Fig.\ref{fig:supp_qual_comp_simple_1}, and Fig.\ref{fig:supp_qual_comp_simple_2}.

While the quantitative results for multi-step instructions are detailed in the main paper, we provide an additional quantitative comparison here using abstract instructions across 12 scenes. Unlike the multi-instruction setting where we evaluated all intermediate edited imagery, here we focus solely on the final output (i.e., the scene state after all edits are completed). We compare our approach against two baselines: BlenderAlchemy~\cite{blenderalchemy} and Blender-MCP~\cite{blendermcp}.
As shown in Tab.~\ref{tab:supp_single_instruction}, our model significantly outperforms all baselines.
% We also provide \textbf{examples of complete MLLM execution runs and sample inputs from our benchmark dataset} in the supplementary materials (see separate files).

% Here we additionally present a quantitative comparison using a dataset of $12$ scenes with general instructions (see ). This evaluation focuses exclusively on the final output (i.e., the scene state after all edits are completed). We compare our approach against two baseline methods: BlenderAlchemy~\cite{blenderalchemy} and Blender-MCP~\cite{blendermcp}. For this experiment, the baseline methods are prompted to execute the general instruction in a single step. As Tab.~\ref{tab:supp_single_instruction} shows, our model outperforms all baselines significantly. Our model also received full plausibility score from all judges in all data.

\input{tables/supp_single_instruction}

\section{Details of the Multi-Tool Library} 
We provide detailed specifications of our multi-tool library, implemented using Blender's Python API (bpy)~\cite{blender}. We start with the visual API functions, followed by our collision-free solver, the constraint loss functions, and finally the collision/floating detector.

\subsection{Visual Probing Functions} The visual probing functions are iteratively called by the \textit{Executor} agent to analyze the scene and select relevant objects or planes for arrangement. The library consists of three primary functions:

\textbf{\textsc{ListObjectsInArea}} returns the names of all objects within a specified image region. This is achieved by rendering an instance map of the scene, extracting all unique IDs from pixels within the input area, and retrieving the object names corresponding to those IDs.

\textbf{\textsc{RayProbe}} returns information regarding the first hitting point of a camera ray given its pixel coordinates. We implement this using bpy's internal ray-casting function to retrieve the position, the surface normal and the object name of the ray's first hit. Additionally, this function extracts the planar surface of the object intersected by the ray. This is accomplished via a breadth-first traversal to identify all neighboring mesh faces of the intersected face that have surface normals within a cosine distance of $0.05$.

\textbf{\textsc{RenderWithHighlight}} generates a rendered image with selected objects highlighted in distinct colors. Similar to the previous functions, this process begins by rendering an instance map. Pixels corresponding to instance IDs in the selection list are then repainted with their assigned colors.

\subsection{Constraint Loss Functions}

In the main paper, we proposed a set of constraints for the solver's input. Here, we detail how each constraint is formulated as a differentiable loss.

\textbf{\textsc{CloseToPix}} penalizes object placements that deviate from the provided pixel coordinates. Given the object's 3D location $\bm{p}$ and the normalized target pixel coordinates $\bm{c}$, the loss is calculated as:
\begin{align}
\mathcal{L}_{c2p} = \lambda_{c2p}\lVert \bm{c} - \mathrm{Proj}_C(\bm{p})\rVert_2 ^2,
\end{align}
where $\mathrm{Proj}_C(\cdot)$ denotes the camera projection function of camera $C$, and the weight parameter $\lambda_{c2p}$ is set to $0.5$.

\textbf{\textsc{Contact}} penalizes placements where the object fails to contact the target plane. Let the object's plane and the target's plane be defined by groups of vertices $\bm{p}_{1,...,n}$ and $\bm{q}_{1,...,m}$, respectively, and the target plane's normal direction denoted as $\bm{n}$. The loss consists of two components: $\mathcal{L}^{touch}_{contact}$, which enforces that the object's plane touches the target plane; and $\mathcal{L}^{above}_{contact}$, which ensures the object's plane resides fully on the outward side of the target plane. Mathematically, let $\bar q = \max_{j=1,...,m}(\bm{q}_j\cdot \bm{n})$ denote the plane's position along the normal direction. The losses are defined as:
\begin{align}
    \mathcal{L}^{touch}_{contact} &= \min_{i=1,...,n}(\lvert \bar q - \bm{p}_i\cdot \bm{n} \rvert_1),  \\
    \mathcal{L}^{above}_{contact} &= \frac{1}{n}\sum_{i=1,...,n}\mathrm{Relu}(\bar q - \bm{p}_i\cdot \bm{n}),  \\
    \mathcal{L}_{contact} &= \lambda^{touch}_{contact}\mathcal{L}^{touch}_{contact} + \lambda^{above}_{contact}\mathcal{L}^{above}_{contact}.
\end{align}
where $\lambda^{touch}_{contact}$ and $\lambda^{above}_{contact}$ are both set to $100$. Note that the contact loss also applies to non-parallel planes.

\textbf{\textsc{NoOverHang}} enforces that the object is placed within the bounds of the target plane. 
While objects can typically be placed fully within the target surface (e.g., a book on a shelf), certain scenarios (e.g., stacking books) may require part of the object to overhang. 
To address this, we developed two modes for this loss: 
\textit{full} mode, which penalizes any part of the object's plane extending beyond the target's convex hull; 
and \textit{center-only} mode, which only evaluates the center of the object's plane.
We first project the vertices of both planes ($\bm{p}_{1,...,n}$ and $\bm{q}_{1,...,m}$) onto the target plane's 2D space (denoted as $\bm{\hat{p}}$ and $\bm{\hat{q}}$). 
We then determine if any $\bm{\hat{p}}$ lies inside the convex hull $\bm{\hat{q}}^{cvx}_{1,...,m'}=\mathrm{ConvexHull}(\bm{\hat{q}}_{1,...,m})$ by applying a cross-product examination with the convex hull's edges $\bm{\hat{e}}_i=\bm{\hat{q}}^{cvx}_{i+1} - \bm{\hat{q}}^{cvx}_{i}$. 
The two modes are defined as:
\begin{align}
\mathcal{L}^{full}_{OH} =& \lambda_{OH}\max_{i,j} (\mathrm{ReLU}((\bm{\hat{p}}_i-\bm{\hat{q}}^{cvx}_j)\times\bm{\hat{e}}_{j})), \\
\mathcal{L}^{center}_{OH} =& \lambda_{OH}\max_{j} (\mathrm{ReLU}((\mathrm{avg}(\bm{\hat{p}})-\bm{\hat{q}}^{cvx}_j)\times\bm{\hat{e}}_{j})) \\
&+ \lambda^{align}_{OH}\lVert\mathrm{avg}(\bm{\hat{p}}) -\mathrm{avg}(\bm{\hat{q}}^{cvx})\rVert,
\end{align}
where $\lambda_{OH}$ and $\lambda^{align}_{OH}$ are set to $20$ and $1$, respectively. 
In the \textit{center-only} mode, an additional $L_2$ loss between the centers of the two planes is applied to encourage placement near the target plane's center.
In practice, the agent selects the mode via an argument in the \textsc{NoOverHang} constraint. 
If \textit{full} mode is selected, only the first formulation is applied. 
Otherwise, the solver first attempts to solve using \textit{full} mode and switches to \textit{center-only} mode if no solution is found.

\textsc{\textbf{Distance}} constrains the distance between two objects to be close to a target value, $dist$. This is implemented as an $L_2$ loss between $dist$ and the Euclidean distance between the centers of the two objects:
\begin{align}
\mathcal{L}_{dist} = \lambda_{dist}\lVert \bm{x}_{o} - \bm{x}_{tgt}\lVert_2^2,
\end{align}
where $\bm{x}_{o}$ and $\bm{x}_{tgt}$ represent the positions of the placed object and the target object, respectively, and $\lambda_{dist}$ is set to $0.3$.

\textsc{\textbf{FaceTo}} enforces that an object faces a specific target object, plane, or camera. Let $\bm{v}_{o}$ denote the object's facing direction and $\bm{v}_{tgt}$ the target facing direction. The loss is calculated as the cosine distance between $\bm{v}_{o}$ and the projection of $\bm{v}_{tgt}$ onto the object's azimuth plane. Letting $\bm{u}_{o}$ be the up vector of the object, the loss is defined as:
\begin{align}
\mathcal{L}_{FT} = \lambda_{FT}\mathrm{CosDist}(\bm{v}_{o}, \bm{v}_{tgt}-\mathrm{SG}((\bm{u}_{o}\cdot\bm{v}_{tgt})\bm{u}_{o})), 
\end{align}
where $\mathrm{SG}$ denotes the stop-gradient function, and $\lambda_{FT}$ is set to $0.5$. Analogously, we implement a \textsc{\textbf{BackTo}} constraint using the same formulation to align the object's back with the target.

\textsc{\textbf{Rotate}} explicitly sets the object's rotation to a specific degree. This is achieved by setting the initial rotation angle of the object to the input degree and fixing it during the optimization process. This constraint is deactivated if any \textsc{FaceTo/BackTo} constraint is present.

\subsection{Collision-Free Solver}
As outlined in Algorithm 1 of the main paper, the solver first perturbs the target pixel coordinates in the \textsc{CloseToPix} constraint with a fixed standard deviation of $0.2$, generating a batch of variant coordinates. It then optimizes a batch of target poses corresponding to each perturbed coordinate.

The positions are initialized at the first intersection point of a ray cast from the pixel coordinates, as determined by the \textsc{RayProbe} function. Initial rotation angles are set to zero, unless a \textsc{Rotate} constraint is explicitly present. We employ the AdamW optimizer for $800$ iterations, with a learning rate initialized at $1e-1$ and linearly decayed to $1e-4$. To simplify the optimization, we fix the rotation along the pitch and roll axes, optimizing only the rotation around the vertical (yaw) axis.

Upon completion, we re-evaluate the loss for each solution in the batch using the original, unperturbed coordinates for the \textsc{CloseToPix} target. Solutions exceeding a loss threshold of $1e-1$ are discarded.

\begin{figure}[t]
  \centering
  \includegraphics[width=0.9\linewidth]{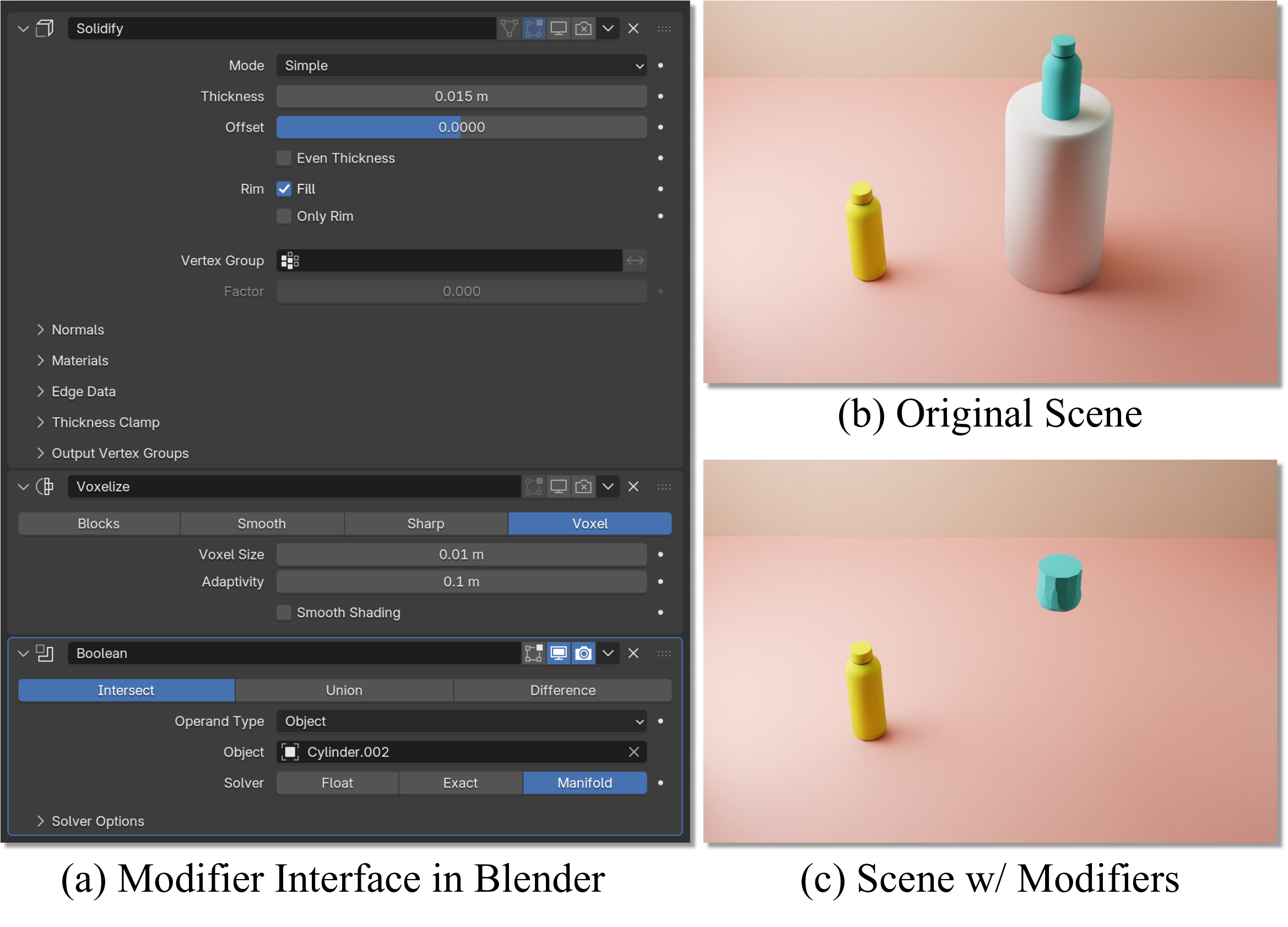}
  \caption{\textbf{Collision Detector.} \textit{(a)} Configuration of our collision detector, consisting of three modifiers with parameter settings as illustrated. \textit{(b-c)} Visualization of the intersection: upon activation, the modifiers extract the collision geometry, i.e. the portion of the bottle below the cylinder's surface.}
  \label{fig:supp_collision_detector}
  \vspace{-15pt}
\end{figure}

Finally, we assess the remaining solutions for collisions, processing them in ascending order of error. Since the standard Blender Python API (bpy) does not expose the native physics engine's collision detection, we leverage geometry modifiers to approximate this. Specifically, we first apply a \textsc{Solidify} modifier to inflate the meshes, ensuring robust detection for thin structures such as walls or floors. Then, we convert the source and target objects into manifold meshes via the \textsc{Remesh} modifier. We then employ a \textsc{Boolean} modifier to compute the intersection volume between objects. A collision is declared if the intersection geometry penetrates the object's bounding box above a threshold of $0.01$. The first solution found to be collision-free is selected as the final output. Figure~\ref{fig:supp_collision_detector} illustrates the modifiers in Blender, and demonstrates how our detector effectively calculates intersection volumes.

\subsection{Floating Detection}
Upon completion of the object arrangement, we conduct a floating check on all objects within the scene. This is implemented by casting a ray from the center of the bottom face of the object's bounding box. If the ray intersects another object within a distance of $0.01$, the object is classified as grounded (i.e., not floating). The validation passes only if all objects that were grounded in the initial state remain grounded in the edited scene.

\section{Prompts for Agents}
\paragraph{\textit{Planner}.} We employ distinct prompting strategies for general instructions and per-step instruction settings. For general instructions, we prompt the planner to autonomously generate a plan based on the current state and historical steps. To prevent infinite loops, we impose a maximum step limit for each task, typically ranging from $3$ to $6$ steps. The model is required to make a feasible plan within this constraint; if the estimated steps exceed this limit, the attempt is automatically classified as a failure. 

\begin{figure*}[t]
  \centering
  \includegraphics[width=\linewidth]{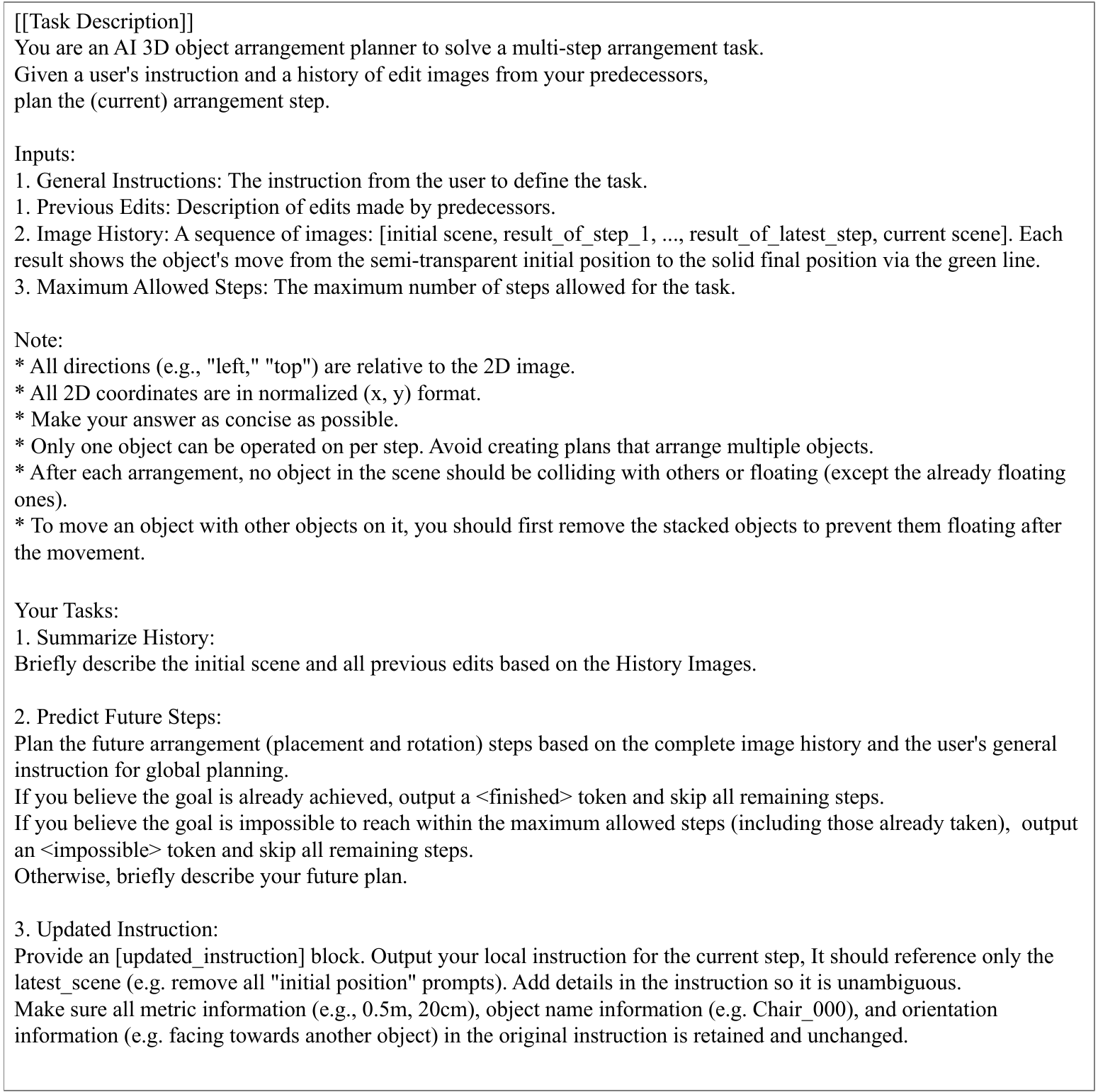}
  \caption{\textbf{Prompt for the \textit{Planner} with General Instructions (Part 1).} }
  \label{fig:supp_planner_prompt_1}
\end{figure*}

\begin{figure*}[t]
  \centering
  \includegraphics[width=\linewidth]{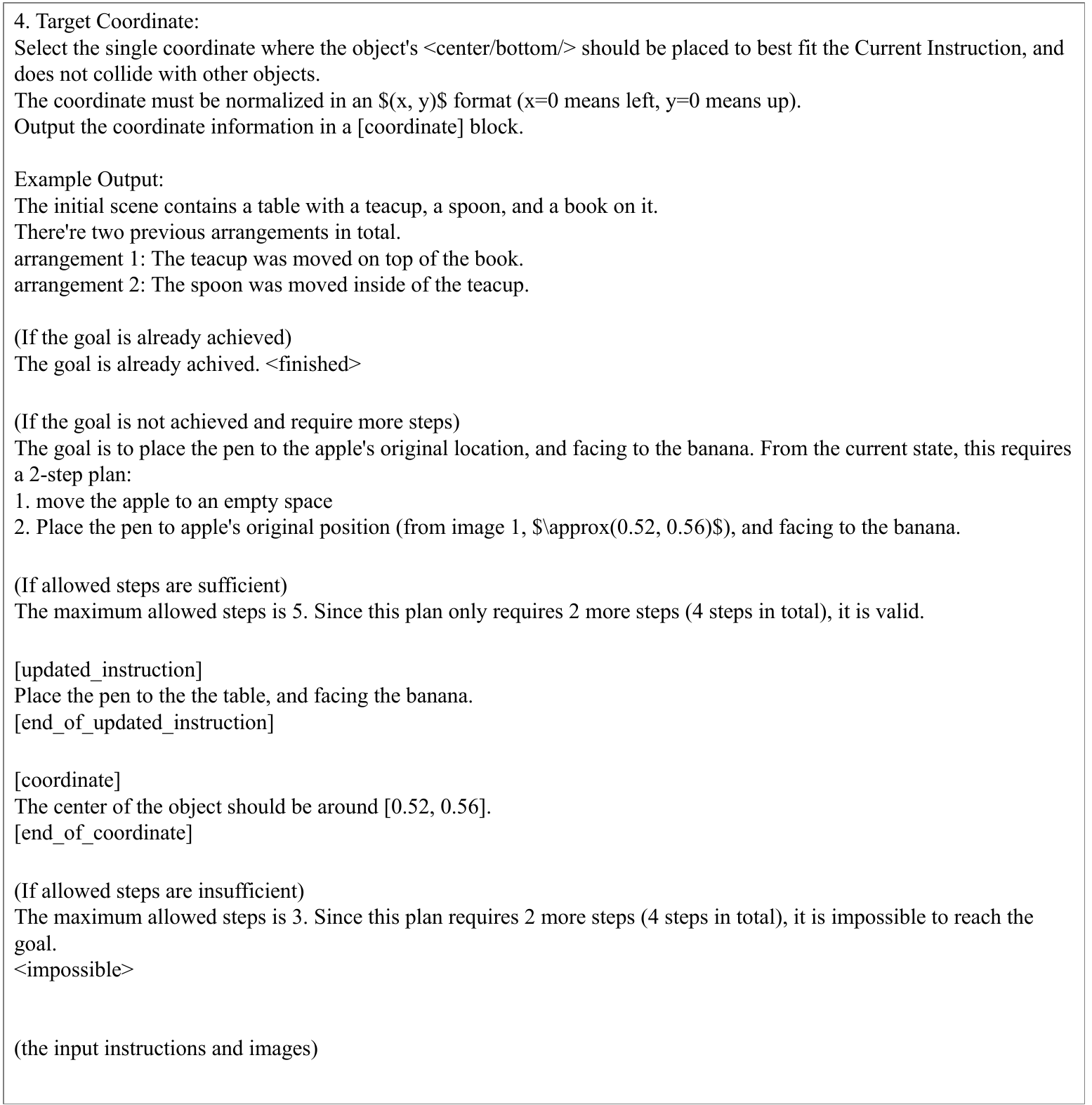}
  \caption{\textbf{Prompt for the \textit{Planner} with General Instructions (Part 2).} }
  \label{fig:supp_planner_prompt_2}
\end{figure*}

\begin{figure*}[t]
  \centering
  \includegraphics[width=\linewidth]{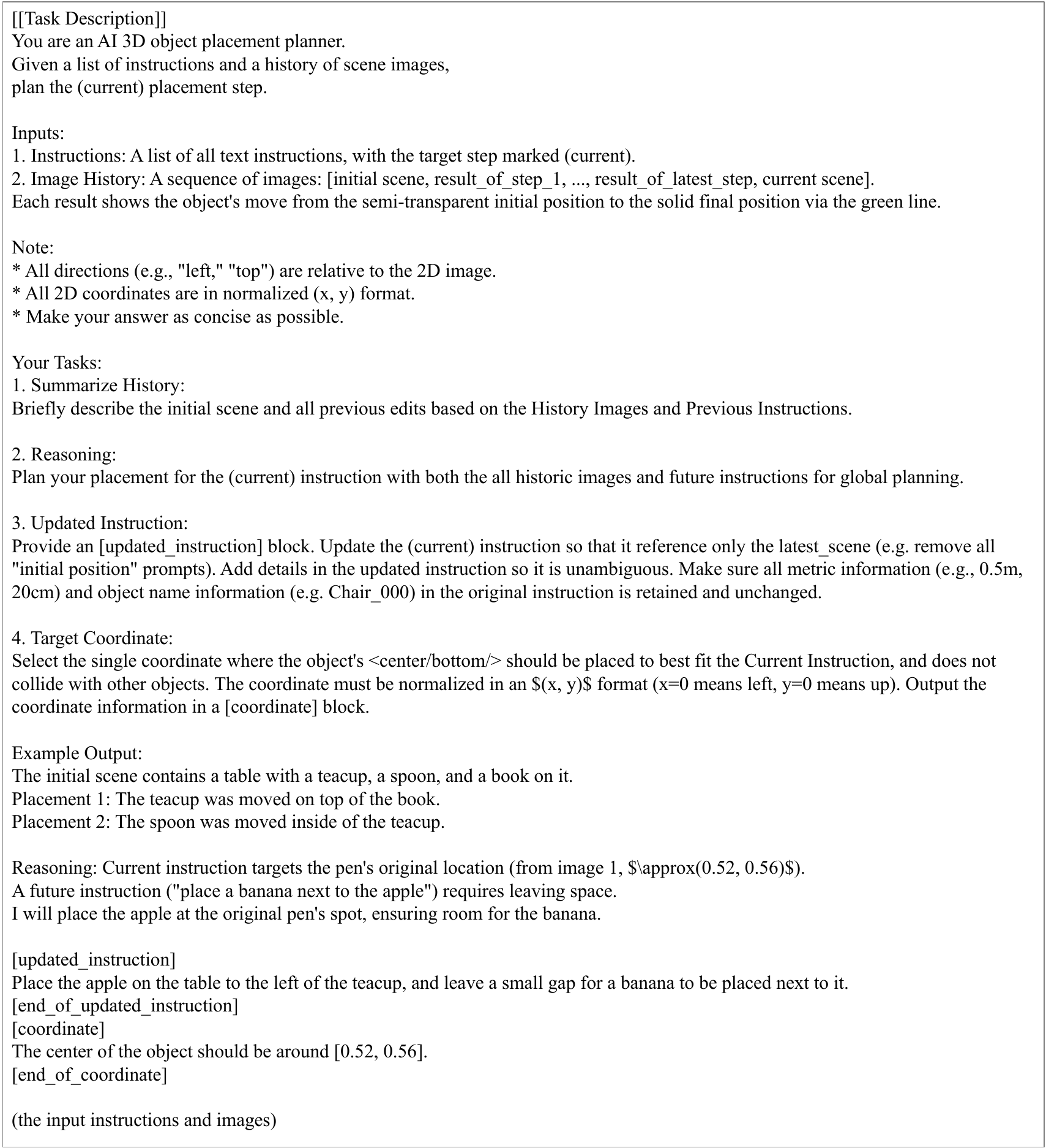}
  \caption{\textbf{Prompt for the \textit{Planner} with Per-step Instructions.} }
  \label{fig:supp_planner_prompt_detailed}
\end{figure*}

In the detailed per-step instruction setting, the model is prompted to read the current instruction, analyze it within the global context (considering both past and future steps), and rewrite it such that the locally-operating \textit{Executor} can execute it based solely on the current state. Figures~\ref{fig:supp_planner_prompt_1} and~\ref{fig:supp_planner_prompt_2} illustrate the prompts for the general instruction setting, while Fig.~\ref{fig:supp_planner_prompt_detailed} shows the prompt for the detailed setting.

\begin{figure*}[t]
  \centering
  \includegraphics[width=\linewidth]{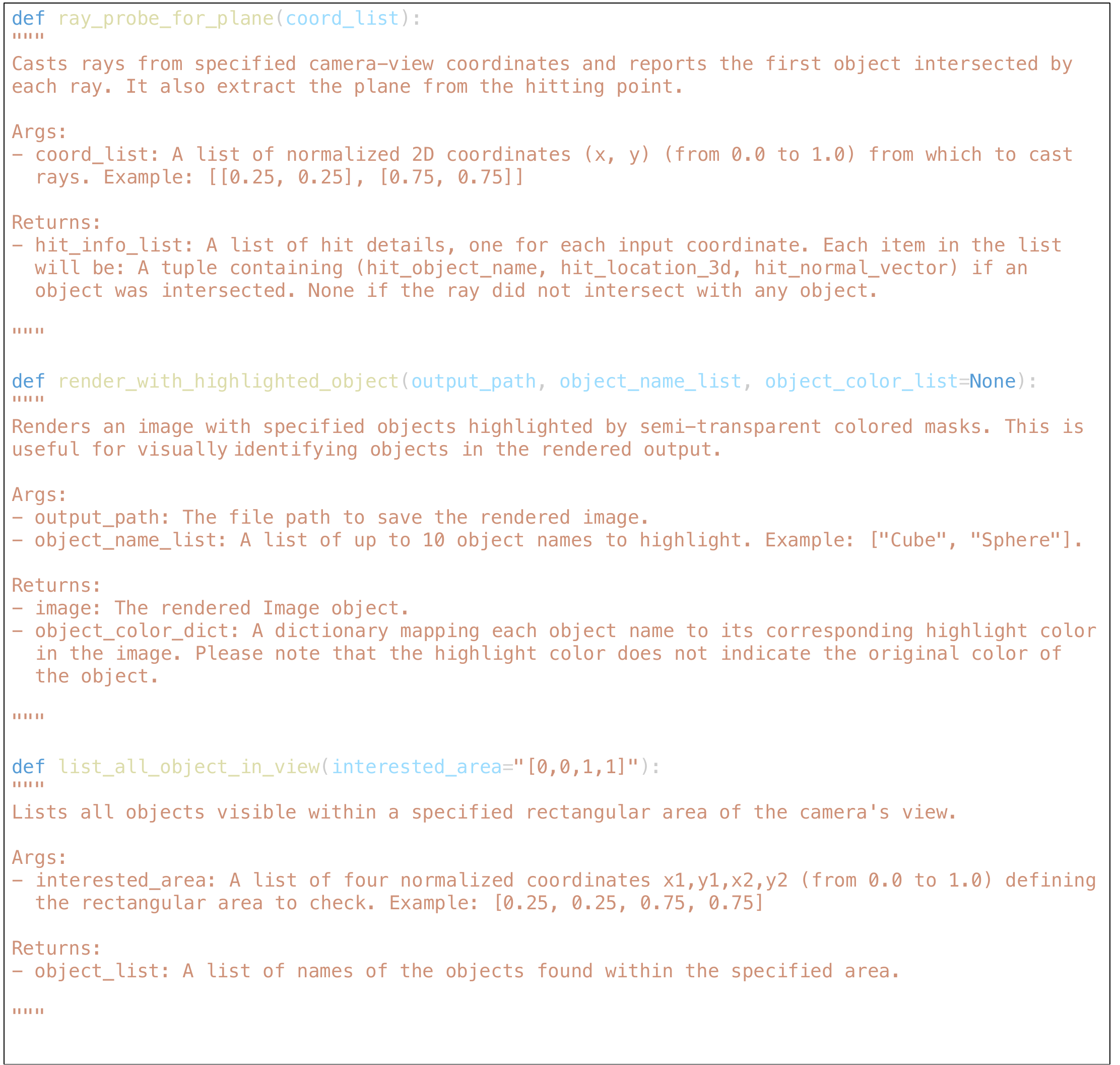}
  \caption{\textbf{Description of Visual API functions.} Note that function names may differ from those in the main text. }
  
  \label{fig:supp_visual_api}
\end{figure*}

\begin{figure*}[t]
  \centering
  \includegraphics[width=\linewidth]{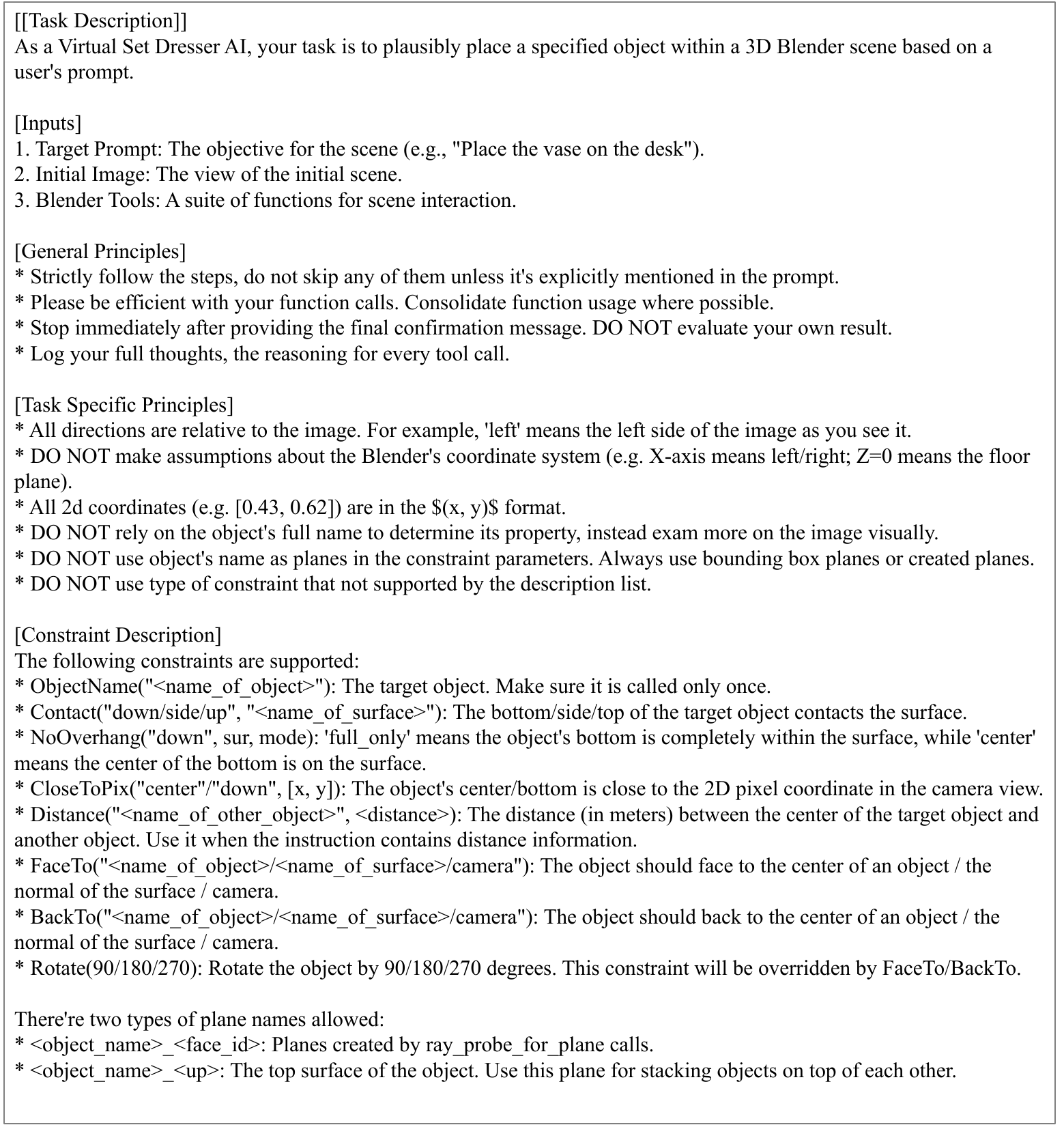}
  \caption{\textbf{Prompt for the \textit{Executor} (Part 1).} }
  \label{fig:supp_executor_prompt_1}
\end{figure*}

\begin{figure*}[t]
  \centering
  \includegraphics[width=\linewidth]{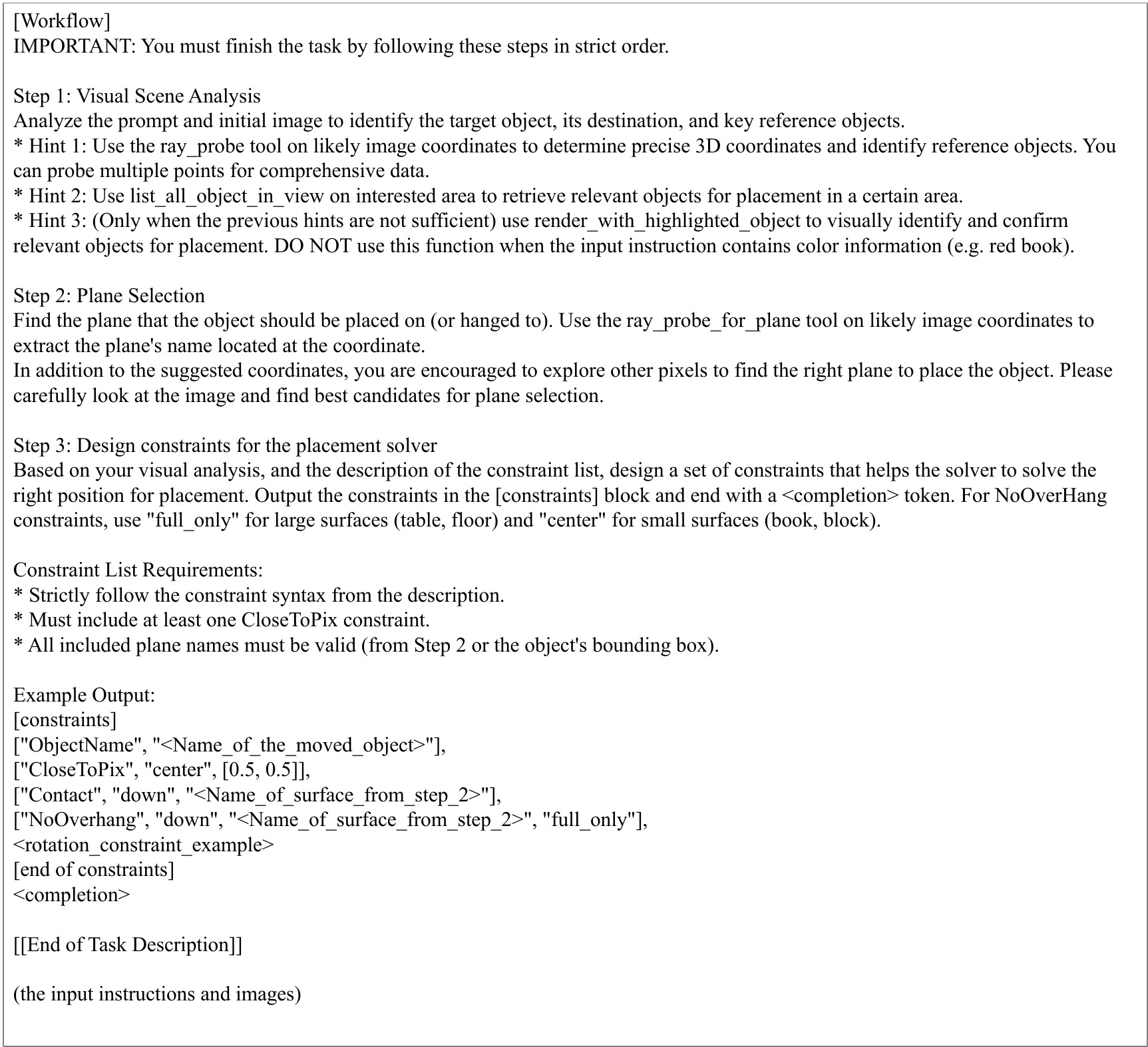}
  \caption{\textbf{Prompt for the \textit{Executor} (Part 2).} }
  \label{fig:supp_executor_prompt_2}
\end{figure*}

\paragraph{\textit{Executor}.} The prompt for the executor is presented in Fig.~\ref{fig:supp_executor_prompt_1} and Fig.~\ref{fig:supp_executor_prompt_2}. Supplementing this, we provide API documentation (as shown in Fig.~\ref{fig:supp_visual_api}) for the visual tools, enabling the model to understand the functionality of each function and the definitions of their parameters.

\begin{figure*}[t]
  \centering
  \includegraphics[width=\linewidth]{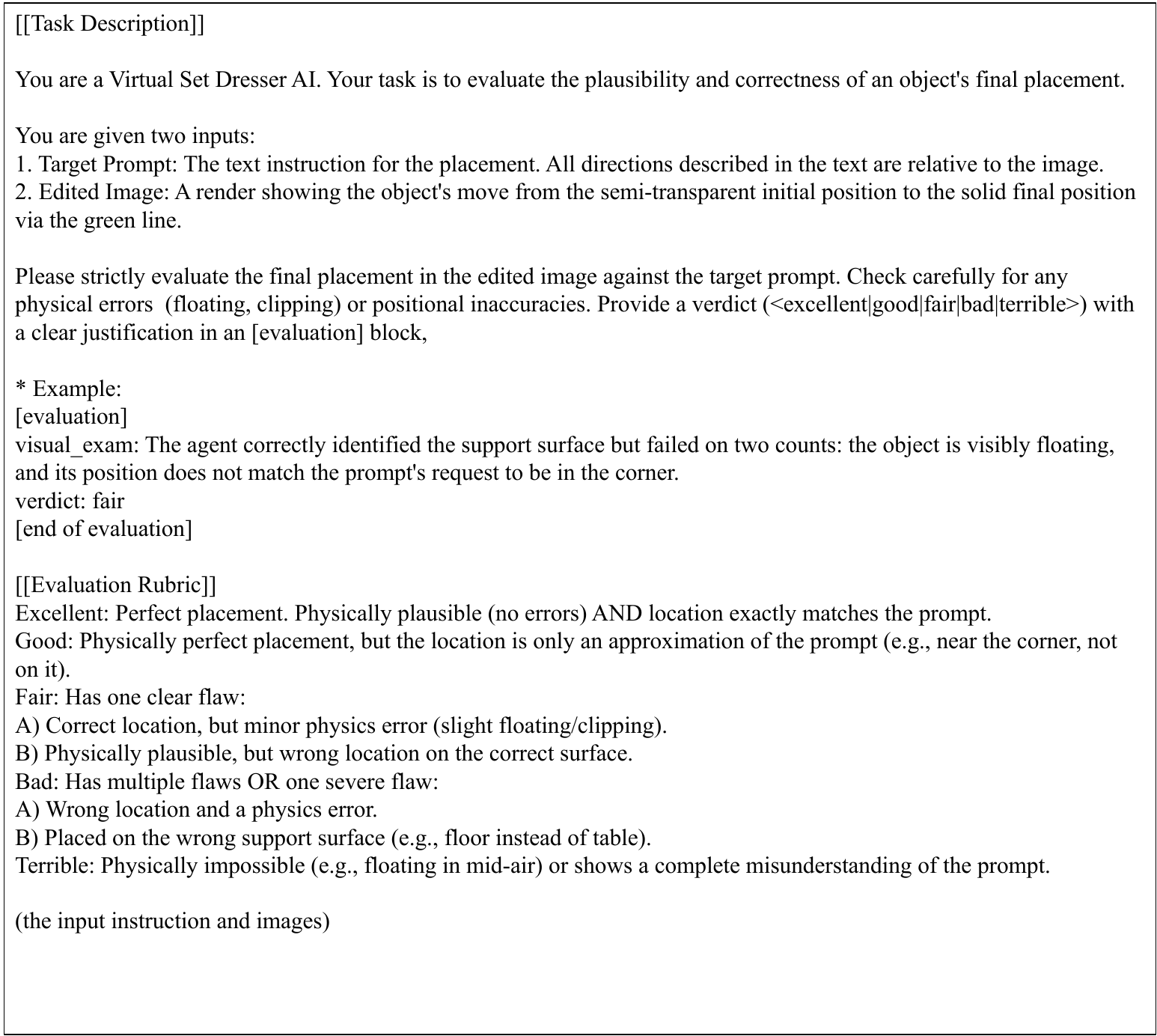}
  \caption{\textbf{Prompt for the \textit{Evaluator}.} }
  \label{fig:supp_evaluator_prompt}
\end{figure*}

\paragraph{\textit{Evaluator}.} Figure~\ref{fig:supp_evaluator_prompt} shows the \textit{Evaluator} prompt. This agent accepts the rendered image of the edited state and the instruction from the \textit{Planner} instruction to assess the edit's plausibility and correctness. In practice, we implement an early termination for efficiency: If all \textit{Evaluators} assign a \textit{good} or \textit{excellent} verdict to the current edit, we accept it as the result for that step and skip remaining attempts.

\begin{figure*}[t]
  \centering
  \includegraphics[width=\linewidth]{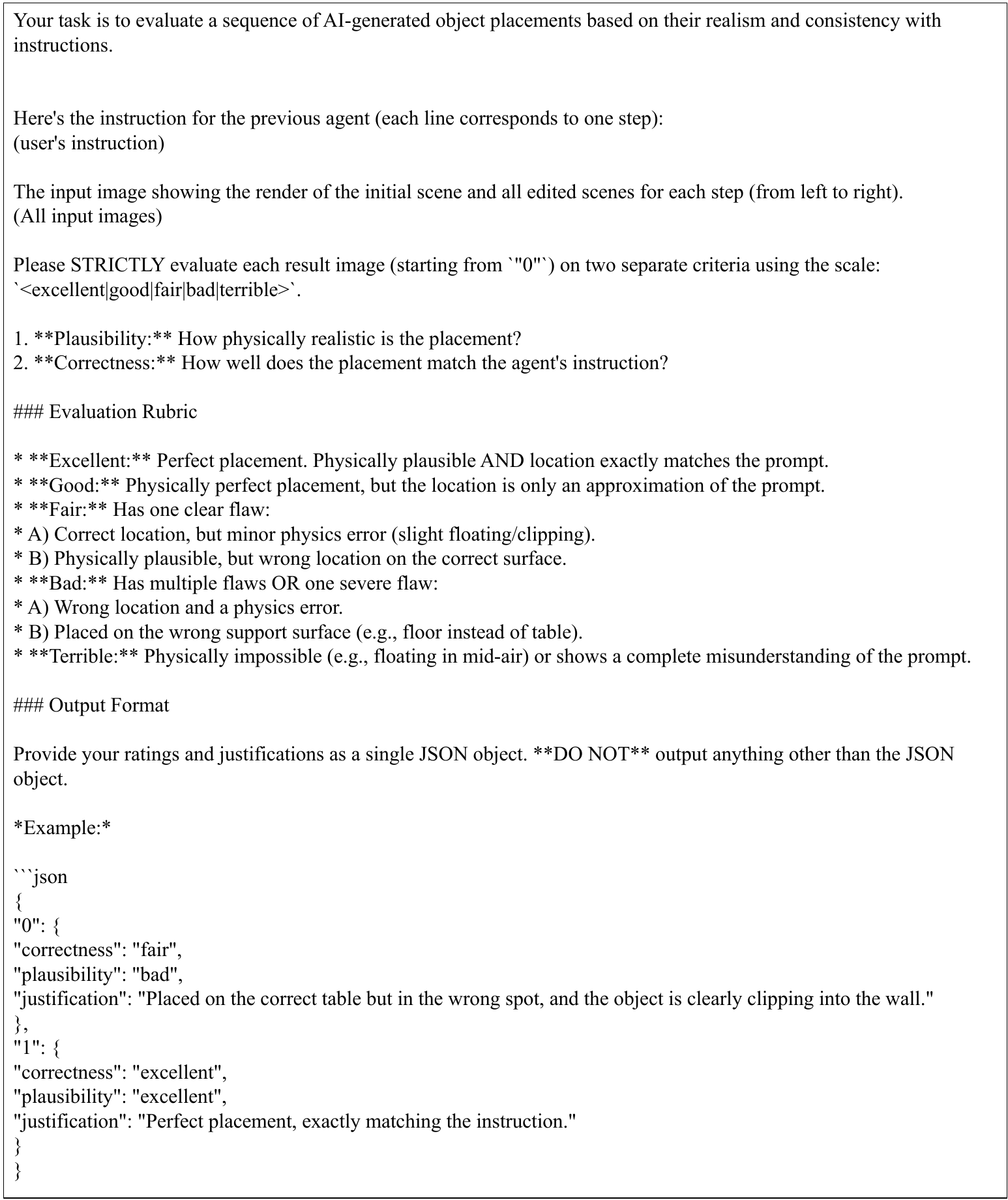}
  \caption{\textbf{Prompt for the Benchmark Judges.} }
  \label{fig:supp_judge_prompt}
\end{figure*}

\paragraph{Benchmark Judges.} In addition to the multi-agent pipeline, we provide the judge prompt used in our benchmark (Fig.~\ref{fig:supp_judge_prompt}), which is adapted from experiments in FirePlace~\cite{fireplace}. This agent examines the results of all steps collectively, evaluating them within a global context to assign scores for each step. For each data sample, we employ $5$ benchmark evaluators and report the average scores.

% \paragraph{\textit{Executor}.} The prompt of the executor is shown in Fig.~\ref{fig:supp_executor_prompt_1} and Fig.~\ref{fig:supp_executor_prompt_2}. Along with it, we also provide the API documentation (As shown in \zhengfei{Fig.}) of the visual tools for the model, so it can understand the functionality of each functions and the meaning of the parameter. 

% \paragraph{\textit{Evaluator}.} The Fig.~\ref{fig:supp_evaluator_prompt} shows the prompt for the evaluator, which takes the render from the edited state and the instruction from the \textit{Planner}, and evaluate the correctness and plausibility of the current operation. 

% \paragraph{Benchmark Judges.} In addition to the multi agent pipeline, we also show the judge prompt used in our benchmark in Fig.~\ref{fig:supp_judge_prompt}, which is adapted from FirePlace's experiment~\cite{fireplace}. Different from the \textit{Evaluator} in our system, this agent looks at result of all steps together, evaluate them with global context, and provides scores for each step's result respectively. For each evaluation data, we employ $5$ benchmark evaluators and take the average scores. 

\section{Adaptive Backtracking}

Algorithm~\ref{alg:supp_plan_search} presents detailed pseudocode for the adaptive backtracking mechanism employed within our search framework.

\RestyleAlgo{ruled}
\SetKwInput{KwInput}{Input}
\SetKwInput{KwParameter}{Parameter}
\SetKwInput{KwOutput}{Output}
\SetKwComment{Comment}{/* }{ */}

\begin{algorithm}[hbt!]
\caption{Plan Search with Adaptive Backtracking $\textsc{Search}$}\label{alg:supp_plan_search}
\KwParameter{Object arrangement pipeline $\mathcal{F}$, maximum allowed steps $s_{max}$, user instruction $T$. }
\KwInput{Previous scenes $\mathcal{S}=\{S_{0,...,k-1}\}$, current step $k$, anchor depth $d_{a}$, current maximum depth $d_{max}$.}
\KwOutput{A sequence of edited scenes $S_{0,...,k}$.}

\Comment{Run model for current step.}
$state, S_k\gets \mathcal{F}$($\mathcal{S}$, $T$, $k$)\;

\uIf{$state=\text{"Complete"}$}{
    \Comment{Search is complete.}
    \Return $\mathcal{S}$\;
}
\uElseIf{$k< s_{max} \And state=\text{"Edited"}$} {
    \Comment{Current step is done.}
    \uIf{$k+1>d_{max}$} {
        \Comment{Reaches a breakthrough. Update anchor depth.}
        $d_{max}, d_a\gets k+1$\;
    }
    $\textsc{Search}(\mathcal{S}\cup\{S_{k}\}, k+1, d_{a}, d_{max})$\;
}
\uElse {
    \Comment{Failed. Backtracking.}
    $d_a\gets d_a/2$\;
    $\textsc{Search}(\{S_{0,...,d_a-1}\}, d_a, d_a, d_{max})$\;
}
\end{algorithm}

\begin{figure}[]
  \centering
  \includegraphics[width=0.97\linewidth]{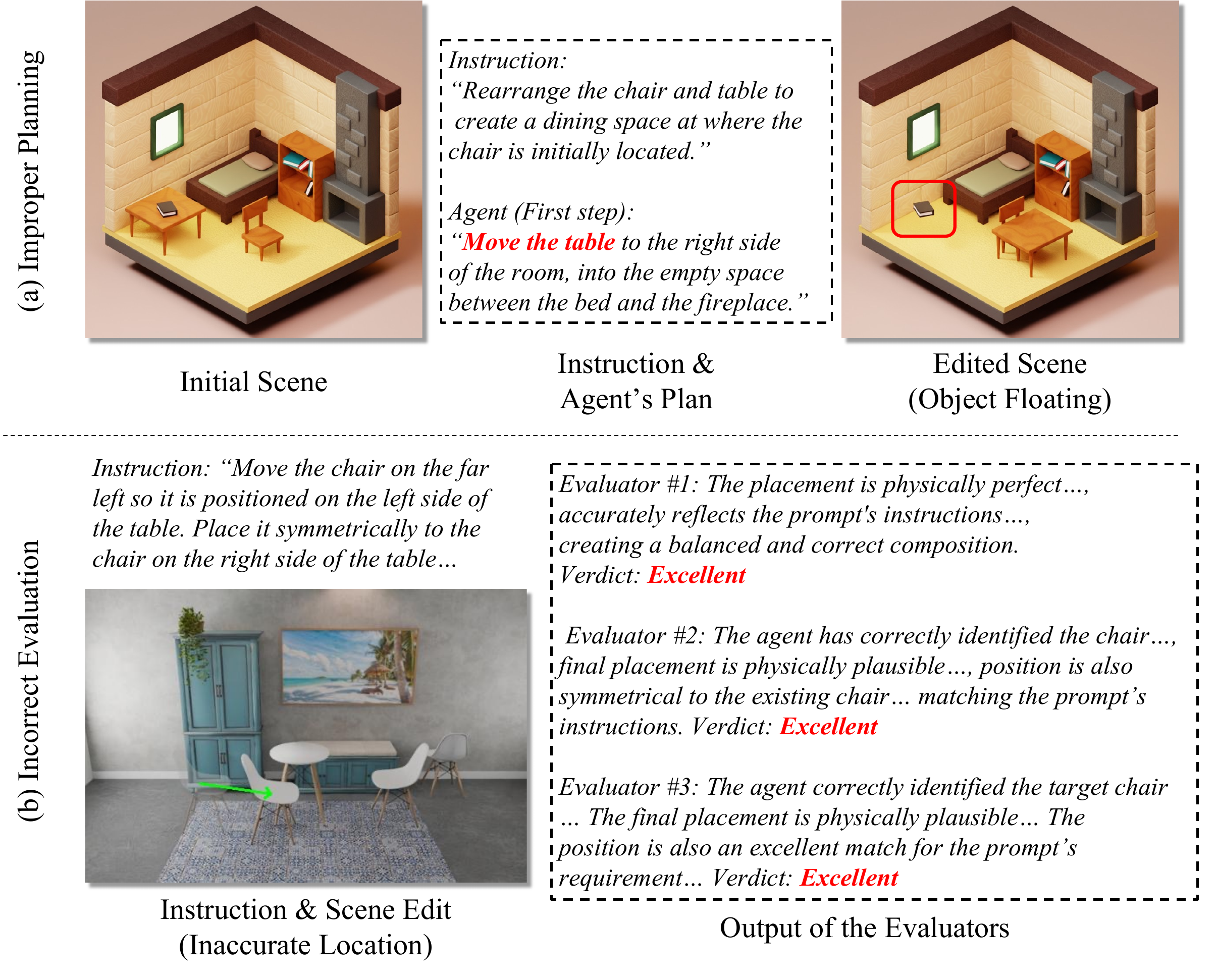}
  \caption{\textbf{Failure Cases.} \textit{Top}: The \textit{Planner} generates an improper plan (i.e., moving the table before clearing it), resulting in floating artifacts. \textit{Bottom}: All three \textit{Evaluators} provide false positive verdicts despite of the arrangement deviating from the instruction. }
  \label{fig:supp_failure_cases}
  \vspace{-14pt}
\end{figure} 

\section{Additional Discussion on Failure Cases} 

Our model significantly enhances the backbone MLLM's capability for iterative object arrangement. However, despite robust design components, the system still can be affected by the inherent limitations of the backbone MLLM, as shown in Fig.~\ref{fig:supp_failure_cases}.

A primary failure mode is planning error. The MLLM may generate an incorrect plan for the current step, resulting in infeasible solutions. While our backtracking algorithm enables recovery from such errors, this trial-and-error process significantly impacts the model's efficiency.

Secondly, as noted in the main paper, the \textit{Evaluator} may occasionally yield false positives (i.e., validating an erroneous arrangement) due to MLLM hallucinations. Although our polling mechanism mitigates this risk, there remains possibilities that the \textit{Evaluators} will reach a consensus on an incorrect verdict. While future advancements in MLLMs may eliminate these issues, we believe that developing more robust evaluation agents for current MLLMs represents a promising direction for future research.

\section{Additional Results} 

We provide supplementary visual results, including qualitative comparisons against baseline methods and extended demonstrations of our model's performance.

\begin{figure*}[t]
  \centering
  \includegraphics[width=\linewidth]{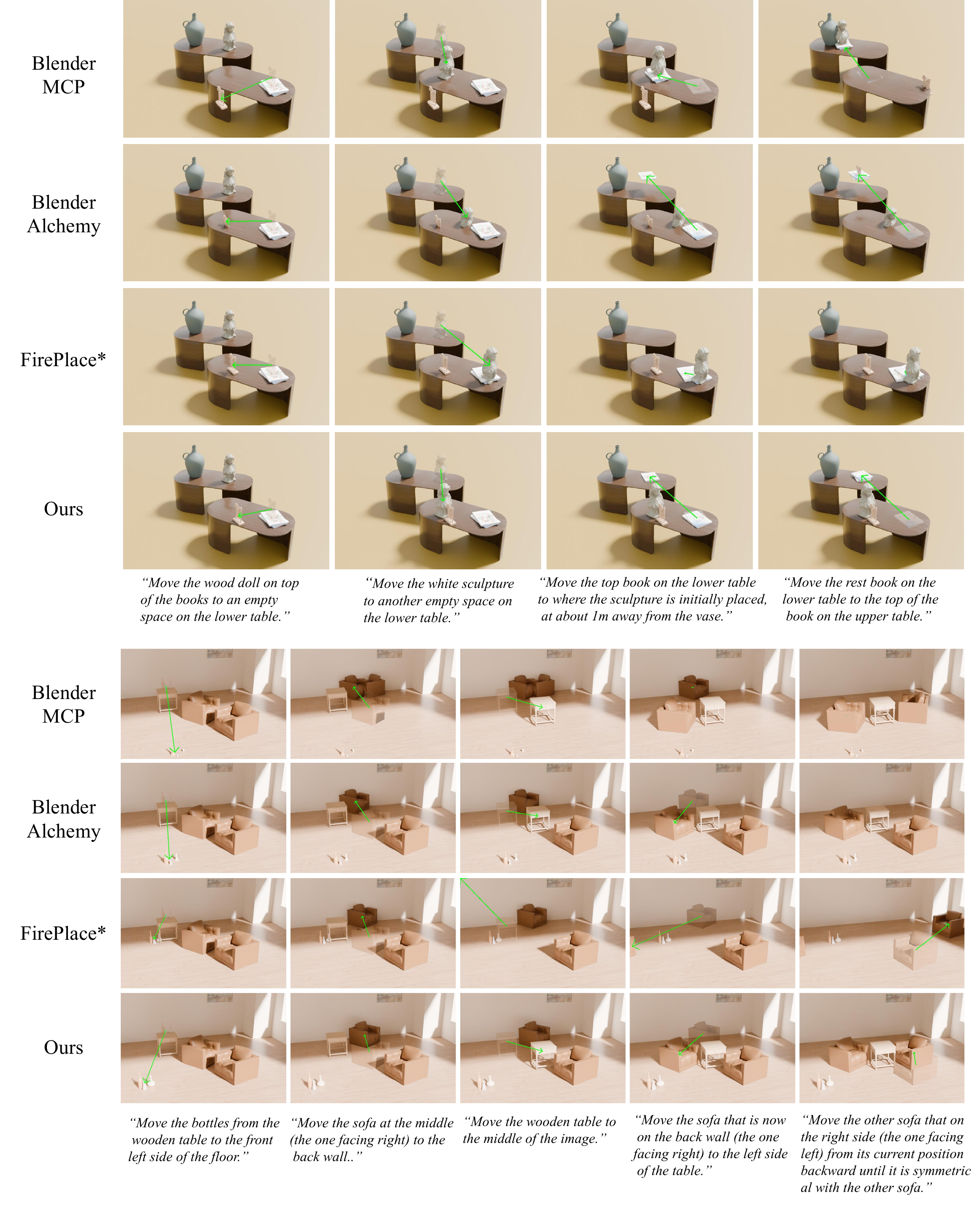}
  \caption{\textbf{Comparison on Per-step Instruction Examples.} The input per-step instructions are shown below the images.}
  \label{fig:supp_qual_comp_per_step_1}
\end{figure*}

\begin{figure*}[t]
  \centering
  \includegraphics[width=0.90\linewidth]{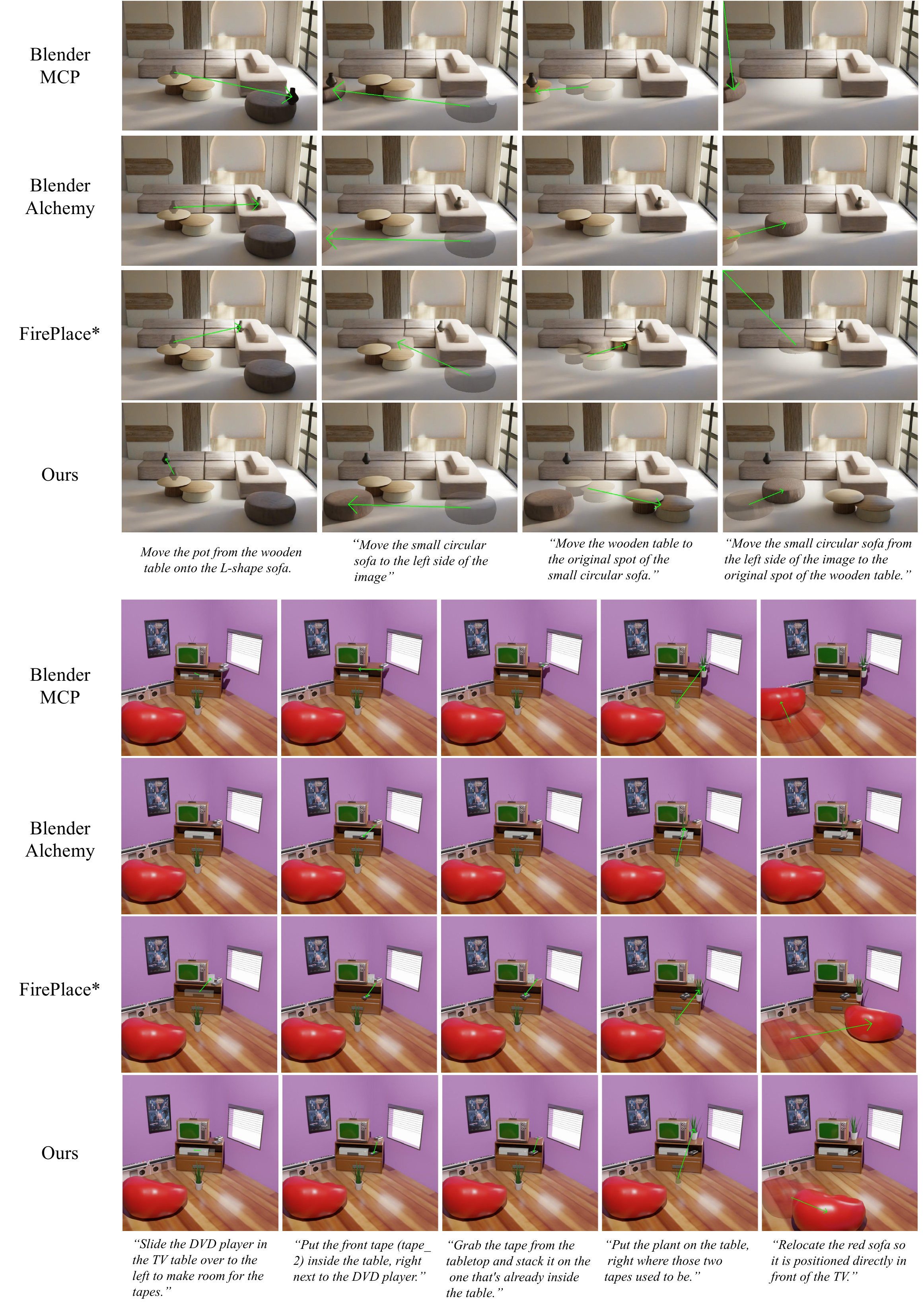}
  \caption{\textbf{Comparison on Per-step Instruction Examples (cont'd).} }
  \label{fig:supp_qual_comp_per_step_2}
\end{figure*}

\begin{figure*}[t]
  \centering
  \includegraphics[width=\linewidth]{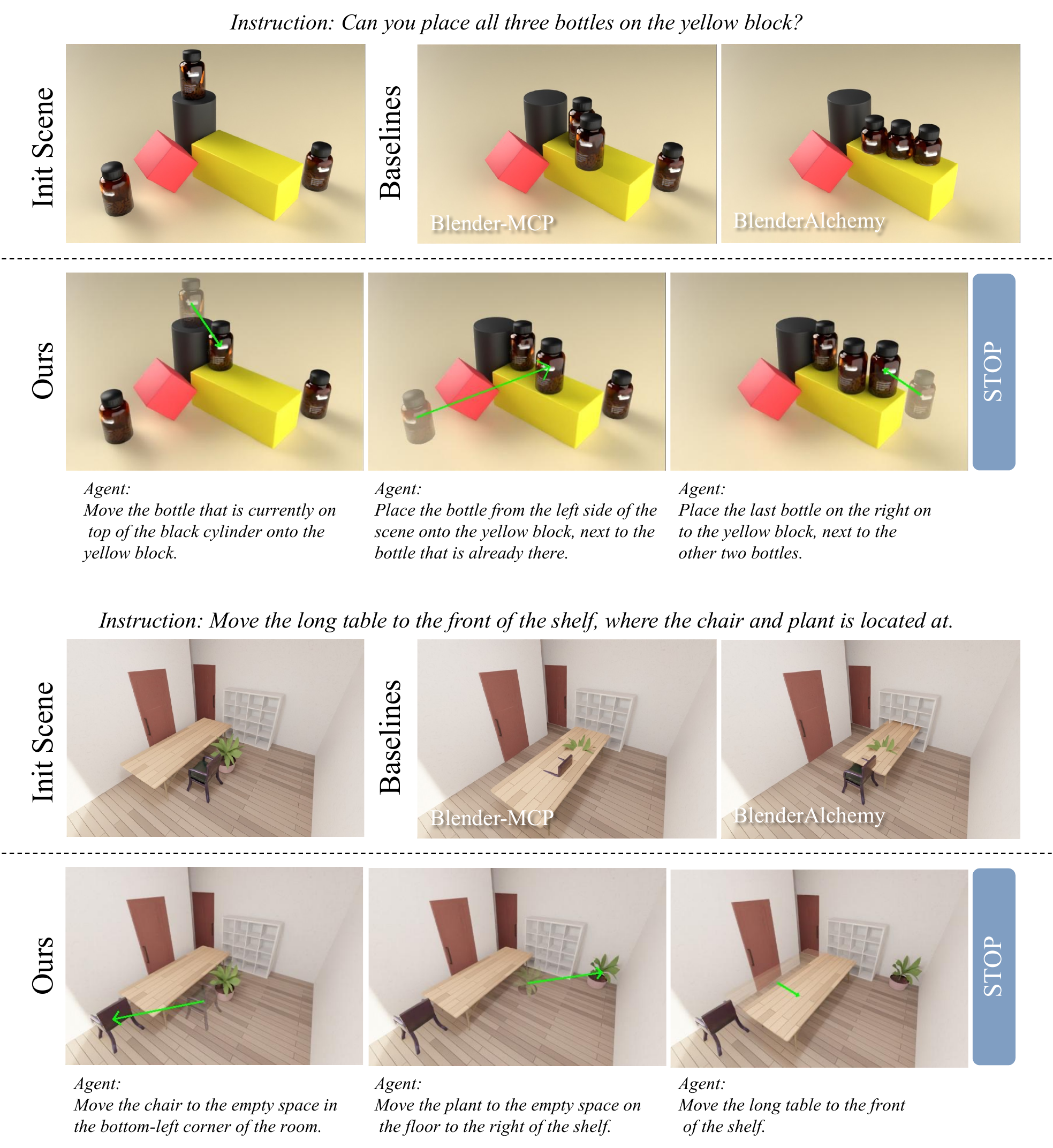}
  \caption{\textbf{Comparison on General Instruction Examples.} The input instruction is shown at the top. We show the initial scene, baseline results, and our model's step-by-step outputs. The plans generated by our model are listed below the edited images.}
  \label{fig:supp_qual_comp_simple_1}
\end{figure*}

\begin{figure*}[t]
  \centering
  \includegraphics[width=0.9\linewidth]{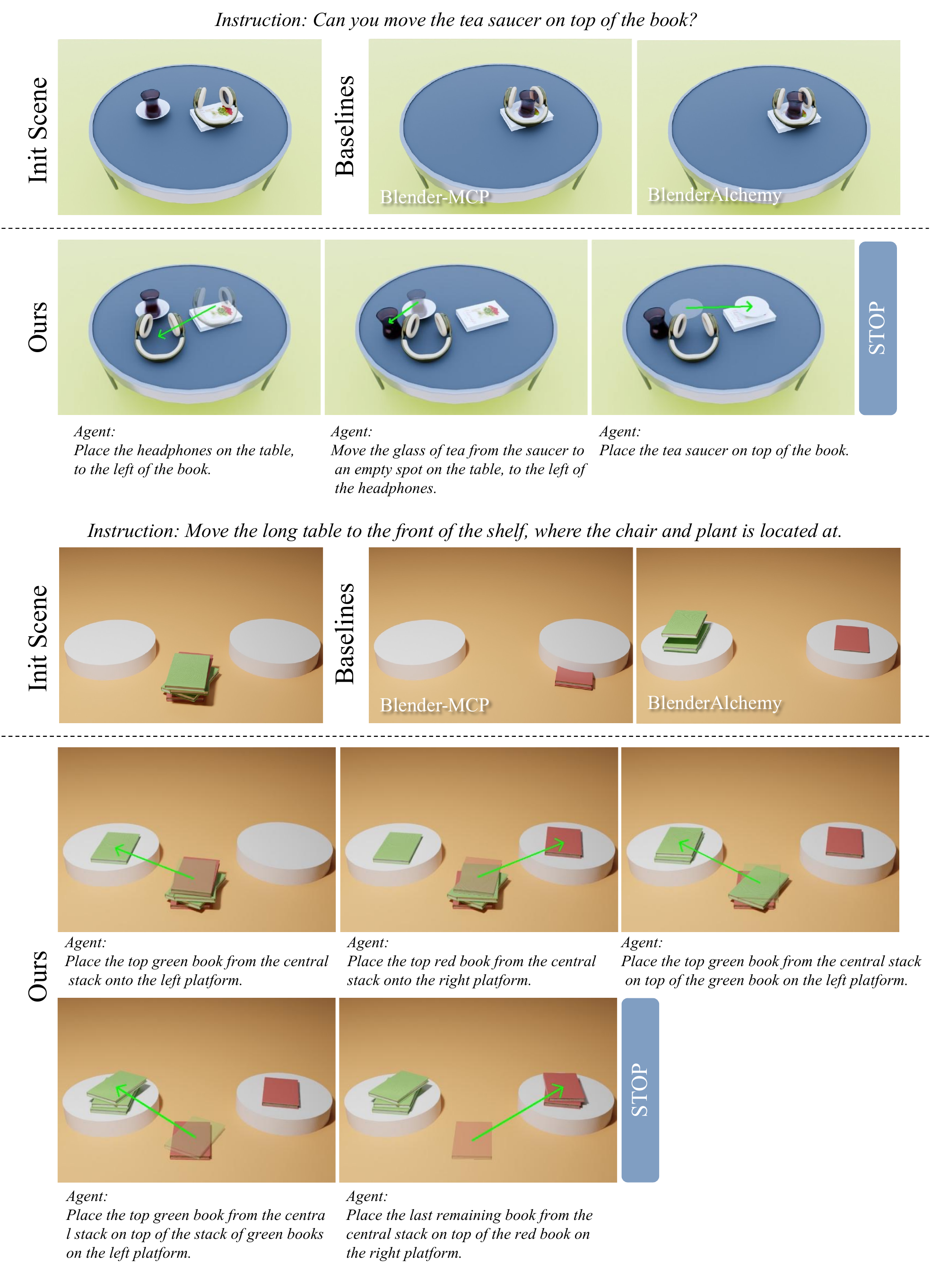}
  \caption{\textbf{Comparison on General Instruction Examples (cont'd).} }
  \label{fig:supp_qual_comp_simple_2}
\end{figure*}

\paragraph{Additional Qualitative Comparisons.} We present further visual comparisons with baseline methods. Figures~\ref{fig:supp_qual_comp_per_step_1} and~\ref{fig:supp_qual_comp_per_step_2} illustrate four examples under the per-step instruction setting, while Figures~\ref{fig:supp_qual_comp_simple_1} and~\ref{fig:supp_qual_comp_simple_2} shows four examples using general instructions.

\begin{figure*}[t]
  \centering
  \includegraphics[width=\linewidth]{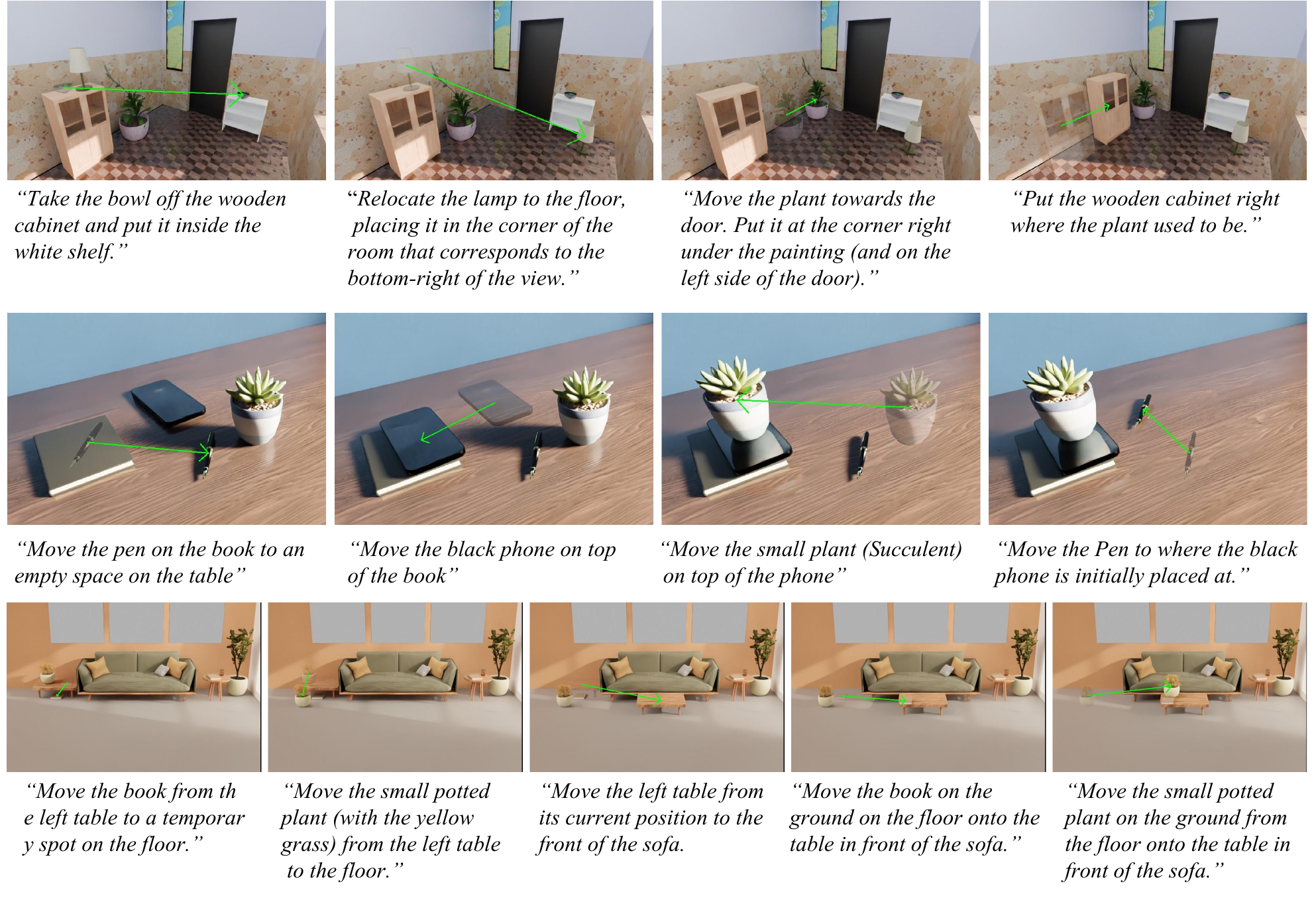}
  \caption{\textbf{Additional Results on Per-step Instruction Examples.} The input instructions are shown below the images.}
  \label{fig:supp_final_per_step}
\end{figure*}

\begin{figure*}[t]
  \centering
  \includegraphics[width=\linewidth]{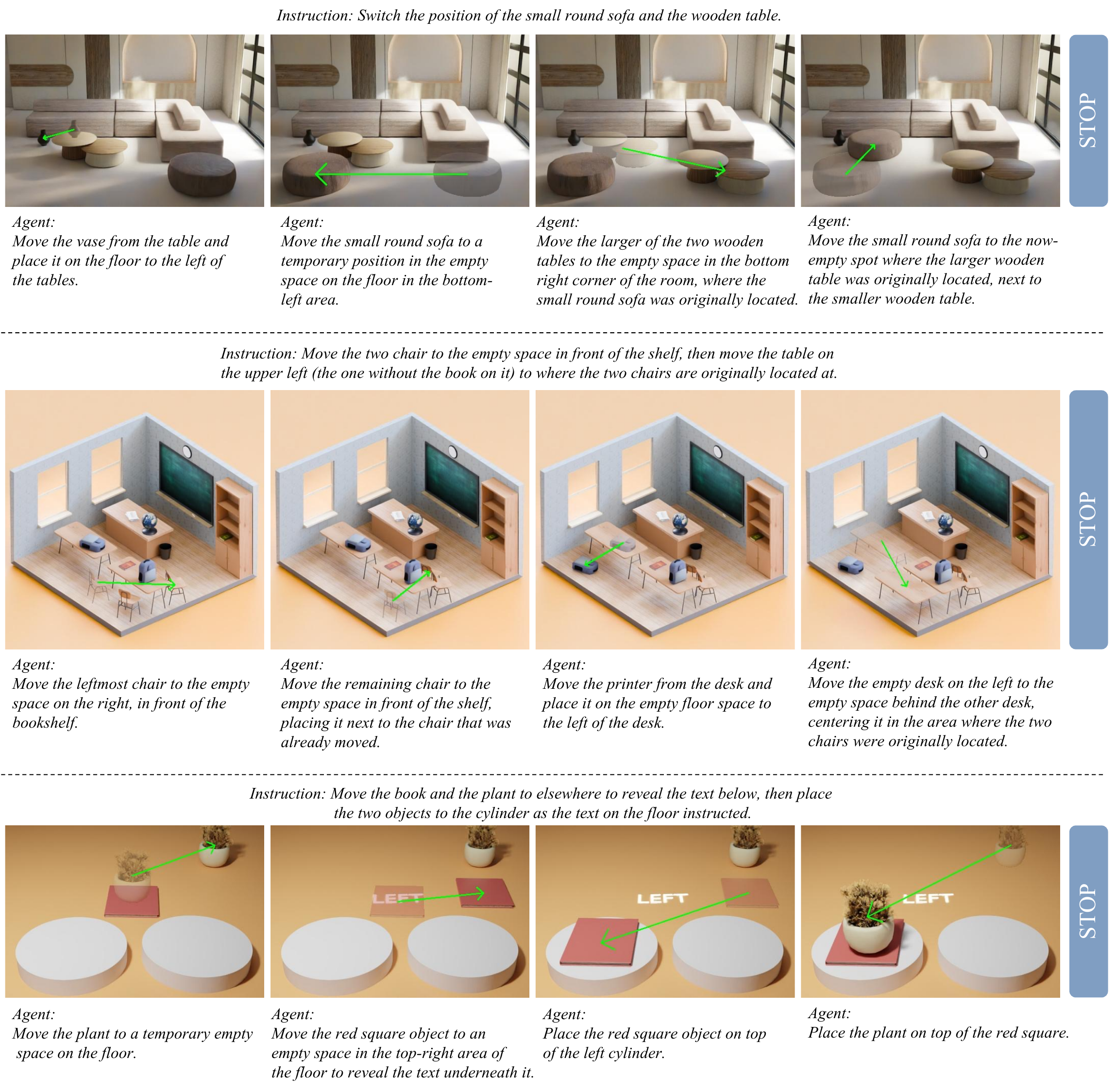}
  \caption{\textbf{Additional Results on Per-Step Instruction Examples.} The input instructions are shown above the images, while the plans generated by the \textit{Planner} agent are shown below.}
  \label{fig:supp_final_simple}
\end{figure*}

\paragraph{Additional Results.} Figures~\ref{fig:supp_final_per_step} and~\ref{fig:supp_final_simple} display additional results generated by our method under per-step and general instruction settings, respectively.

\clearpage

%%%%%%%%%%%%%%%
% Comparisons
%%%%%%%%%%%%%%%

%%%%%%%%%%%%%%%
% More Results
%%%%%%%%%%%%%%%

%%%%%%%%%%%%%%%
% Prompts
%%%%%%%%%%%%%%%


%% file: tables/supp_single_instruction.tex
\begin{table}[h]
\centering
\setlength{\tabcolsep}{2.5pt}
\renewcommand{\arraystretch}{1.4}
\small
\begin{tabular}{lcccc}
\toprule
Method & Coll.\%$\downarrow$ & Fl.\%$\downarrow$ &  Plaus.$\uparrow$ & Const.$\uparrow$ \\
\midrule
% w/o Multi-Tool Library & 0.495 & 0.711 & 3.484 & 3.103 \\
% w/o Backtracking & 0.036 & 0.054 & 3.703 & 3.549 \\
% Single Agent & \textbf{0.000} & \textbf{0.000} & 3.623 & 3.328 \\
% w/o Visual Annotation & \textbf{0.000} & \textbf{0.000} & 3.755 & \textbf{3.638} \\
BlenderAlchemy~\cite{blenderalchemy} & 0.500 & 0.333 & 3.433 & 2.833 \\
Blender-MCP~\cite{blendermcp} & 0.416 & 0.333 & 3.800 & 2.517 \\
\midrule
Ours & \textbf{0.000} & \textbf{0.000} & \textbf{4.000} & \textbf{3.383} \\
\bottomrule
\end{tabular}
\caption{\textbf{Evaluation on General Instructions.} Our model outperforms baselines and produces artifact-free outputs.}
\label{tab:supp_single_instruction}
\vspace{-10pt}
\end{table}